\let\clineorig\cline
\documentclass[sn-mathphys,Numbered,iicol]{sn-jnl}

\usepackage{lineno}

\usepackage{graphicx}%
\usepackage{multirow}%
\usepackage{amsmath,amssymb,amsfonts}%
\usepackage{amsthm}%
\usepackage{mathrsfs}%
\usepackage[title]{appendix}%
\usepackage{xcolor}%
\usepackage{textcomp}%
\usepackage{manyfoot}%
\usepackage{booktabs}%
\usepackage{algorithm}%
\usepackage{algorithmicx}%
\usepackage{algpseudocode}%
\usepackage{listings}%

\theoremstyle{thmstyleone}%
\theoremstyle{thmstyletwo}%

\theoremstyle{thmstylethree}%

\usepackage{amsmath,amsfonts}

\usepackage{graphicx}
\usepackage{textcomp}
\usepackage{xcolor}

\usepackage{algorithm}
\usepackage{algpseudocode}
\usepackage{bm}
\usepackage{float}
\usepackage{subfigure}
\usepackage{graphicx}
\usepackage{color}
\usepackage{soul}
\usepackage{enumitem}
\usepackage{wrapfig}
\usepackage{arydshln}
\usepackage{soul, color, xcolor}

\algnewcommand\algorithmicparam{\textbf{Parameter:}}
\algnewcommand\Param{\item[\algorithmicparam]}

\begin{document}

\title[DTA]{DTA: Distribution Transform-based Attack for Query-Limited Scenario}

\author[1]{\fnm{Renyang} \sur{Liu}}\email{ryliu@mail.ynu.edu.cn}
\equalcont{This paper was accepted by Cybersecurity.}

\author[1]{\fnm{Wei} \sur{Zhou}}\email{zwei@ynu.edu.cn}
\author[1]{\fnm{Xin} \sur{Jin}}\email{xinjin@ynu.edu.cn}
\author[1]{\fnm{Song} \sur{Gao}}\email{gaos@ynu.edu.cn}

\author[3]{\fnm{Yuanyu} \sur{Wang}}\email{wxyjin232425@163.com}


\author*[2]{\fnm{Ruxin} \sur{Wang}}\email{rosinwang@gmail.com}

\affil[1]{\orgname{Yunnan University}, \orgaddress{ \city{Kunming}, \postcode{650504}, \state{Yunnan}, \country{China}}}

\affil[2]{\orgname{Alibaba Group}, \orgaddress{ \city{Beijing}, \postcode{100102}, \country{China}}}


\affil[3]{\orgname{Kunming Institute of Physics}, \orgaddress{ \city{Kunming}, \postcode{650221}, \state{Yunnan}, \country{China}}}


\abstract{In generating adversarial examples, the conventional black-box attack methods rely on sufficient feedback from the to-be-attacked models by repeatedly querying until the attack is successful, which usually results in thousands of trials during an attack. This may be unacceptable in real applications since Machine Learning as a Service Platform (MLaaS) usually only returns the final result (i.e., hard-label) to the client and a system equipped with certain defense mechanisms could easily detect malicious queries. By contrast, a feasible way is a hard-label attack that simulates an attacked action being permitted to conduct a limited number of queries. To implement this idea, in this paper, we bypass the dependency on the to-be-attacked model and benefit from the characteristics of the distributions of adversarial examples to reformulate the attack problem in a distribution transform manner and propose a distribution transform-based attack (DTA). DTA builds a statistical mapping from the benign example to its adversarial counterparts by tackling the conditional likelihood under the hard-label black-box settings. In this way, it is no longer necessary to query the target model frequently. A well-trained DTA model can directly and efficiently generate a batch of adversarial examples for a certain input, which can be used to attack un-seen models based on the assumed transferability. Furthermore, we surprisingly find that the well-trained DTA model is not sensitive to the semantic spaces of the training dataset, meaning that the model yields acceptable attack performance on other datasets. Extensive experiments validate the effectiveness of the proposed idea and the superiority of DTA over the state-of-the-art.}


\maketitle

\section{Introduction}
\vspace{-7pt}
The recent progress in machine learning reveals a critical problem of deep neural networks (DNNs), which states that most of DNNs are vulnerable to adversarial examples, i.e., being misled by particular examples corrupted by human imperceptible noise \cite{corr/SzegedyZSBEGF13,corr/GoodfellowSS14,iclr/KurakinGB17a,cvpr/DongLPS0HL18}. Such an unrobust property and its inexplicability have attracted extensive research attention that was devoted to improving the model's robustness and AI security. While most of the existing studies focus on adversarial attacks in a synthesizing way, i.e., the adversarial examples are generated by directly modifying the pixels of digital images, certain trials have shown that it is possible to attack an AI system physically \cite{cvpr/DuanM00QY20,liu2022threats}. A typical scenario is autonomous driving, where the driving system relies on deep learning-based techniques to identify traffic signs or other road information for accurate driving decisions. The studies by Liu et al. \cite{AAAI/LiuLFMZXT19} and Eykholt et al. \cite{cvpr/EykholtEF0RXPKS18} show that the well-designed disturbances imposed on the traffic signs can easily deceive the recognition module in the driving system, bringing a significant threat to people lives and properties. 

While various defense methods for adversarial attacks are constantly being proposed \cite{cvpr/AkhtarLM18,wang2021lsgan,icml/MadaanSH20,icml/ZhangLLT20,guo2023towards}, more powerful attack methods
\cite{sp/Carlini017,sun2018survey,mirsky2023ipatch} are emerging increasingly and have been able to fight against those defense methods. This attack-defense game will continue along with the development of deep learning and modern AI systems.

The literature on adversarial attacks can be grouped into two classes: white-box and black-box attacks. The white-box attack conveys the case that the details of the target model, such as the structure and the parameters, are known before designing the attack method. In contrast, the black-box setting states that the model details are inaccessible, but only the hard-label or the label probability returned by the target model concerning a specific input can be obtained via a querying-based attack. Clearly, the black-box attack is more feasible than the white-box attack in real applications since the technical details of an online artificial intelligence system are invisible to the public in a general sense, especially for the hard-label setting.

A typical option for the attackers in the black-box setting is to use thousands of queries to collect enough feedback for optimizing the adversarial example iteratively, which is called the optimization-based attack. Nevertheless, the problem here is that the querying-and-optimizing process could result in massive consumption of computing resources and time \cite{icml/GuoGYWW19,aaai/TuTC0ZYHC19}, which results in an inefficient way to perform a successful attack. On the other hand, an advanced AI system could be equipped with certain defense mechanisms that resist intentional attacks \cite{interspeech/WuLL20}, a good case in point is the Google Cloud Vision API (GCV)\footnote{https://cloud.google.com/vision}. In this case, too many trials of attacking can be easily detected by the system. Hence, all these conditions dramatically limit the applicability of the black-box attack and pose the necessity of a query-limited hard-label attack strategy in practical applications. 

Besides, the optimization-based attack methods target overly fitting the adversarial examples on the target model so as to achieve a high attack performance. We empirically find that those generated adversarial examples have low transferability and cannot attack other target models effectively. This limits the possibility of exploring the cross-model knowledge about adversarial robustness.

To solve the problems discussed above, in this paper, we formulate the synthesis of adversarial examples in a distribution transform manner. We advocate that the adversarial distribution and the normal distribution are misaligned but transferable. The common parts during transfer can be well-conditioned on the input example itself. The misaligned parts are enriched by employing a generative model that recovers the distributions based on random noise and conditions. Assuming that the vulnerability of different deep models exhibits similar effects, we can reasonably collect many adversarial examples from the existing attack methods, which are then used to characterize the adversarial distribution. In this way, it is possible to optimize the generative model that synthesizes the adversarial examples in a statistical pipeline. As a result, the model can generate batches of examples for attacking without too many queries. To be clear, this advantage benefits from the distribution of the existing adversarial examples and their transferability, which are encoded by the generative model. It is also allowed to apply the attack model on a different data source that has not been involved in training. In the implementation, we develop a conditional normalizing flow-based model to achieve the above goal. The main contributions of this paper can be summarized as follows:

\begin{itemize}[leftmargin=*, itemsep=1pt, topsep=1pt, parsep=0pt]
    \item We formulate the black-box attack problem as a generative framework from the perspective that the adversarial distribution can be translated from the normal distribution under certain conditions. Within this perspective, the adversarial examples are transferable across different models and different image contents.
    
    \item We develop a conditional normalizing flow-based attack method (DTA) that simulates the transformation from the normal distribution to the adversarial distribution. Unlike the existing black-box methods, which need thousands of queries, DTA significantly reduces the query times during an attack while achieving an acceptable attack success rate. Notably, DTA requires only ONE query to perform a successful attack in most cases.
    
    \item The proposed DTA can generate adversarial examples with high transferability to different black-box models. The well-trained model is not sensitive to the semantic spaces of the training dataset, and we empirically demonstrate that the model trained on ImageNet can be used to generate effective adversarial examples on other datasets.
    
    \item Extensive evaluations on black-box attacks show that the proposed DTA beats the state-of-the-art hard-label attacks in the aspects of attack success rate, query times and transferability, which demonstrate the validity of the proposed DTA in the adversarial attack.
\end{itemize}

The rest of this paper is organized as follows. We briefly review the methods relating to adversarial attacks in Sec. \ref{Sec:related}. In Sec. \ref{sec:preliminary}, we provide the preliminaries of adversarial attack and normalizing flow. Sec. \ref{sec:methodology} introduces the details of the proposed DTA framework. The experiments are presented in Sec. \ref{Sec:experiments}, with the conclusion drawn in Sec. \ref{Sec:conclusion}.

\section{Related Work}
\label{Sec:related}
In this section, we briefly review the most relevant methods to the current work. For comprehensive literature on adversarial attacks (including white-box and black-box), please refer to \cite{ica3pp/DingX20,corr/abs-1810-00069}.

\subsection{Black-box Attack}
A typical case of adversarial attack is the black-box setting that concerns the practice in real applications. Due to the limited information about the target model, the black-box attack is more difficult than the white-box one and receives limited attention from the community. The rationale of most existing methods is the transferability of the adversarial examples across models, which allows the examples generated using the white-box methods to attack the black-box models. For example, the integrated adversarial training method proposed by Tramer et al. \cite{iclr/TramerKPGBM18} and the image transformation method proposed by Guo et al. \cite{iclr/GuoRCM18} could effectively carry out the transfer attack.

The ZOO attack proposed by Chen et al. \cite{ccs/ChenZSYH17} was one of the earliest black-box attack methods based on queries, which employed the zero-order optimization to construct a zero-step estimator by querying a target model and then, used the estimated gradient to minimize the Carlini and Wagner (C\&W) loss \cite{sp/Carlini017} to find adversarial examples. Ilyas et al. \cite{icml/IlyasEAL18} employed the normal distribution search density to estimate the gradient of the DNN classifier $ F(x) $ and adopted the projected gradient descent method to minimize the loss of generating adversarial examples. Instead of minimizing the target of adversarial example generation, $\mathcal{N}$ ATTACK \cite{icml/LiLWZG19} tried to fit the distribution around the clean data, which was followed by the adversarial examples. In another work, Ilias et al. \cite{iclr/IlyasEM19} observed that the gradient used by PGD showed a high correlation in time and data and then, used the slot machine optimization techniques to integrate the prior knowledge about gradients into the attack, thus proposing a method called Bandits \& Priors which reduced the number of queries during an attack.

\subsection{Adversarial Attacks using Generative Models}
The existing adversarial attack methods based on generative models generally rely on the generative adversarial network (GAN), which is used to synthesize adversarial examples \cite{aaai/BalujaF18,iclr/WangY19,iclr/Huang020}. Most of these methods focus on the white-box attack, where the gradient of the target model is required to update the parameters of GAN. In the black-box setting, a surrogate model is used to approximate the output of the target model, which also drives the gradients of the former to approximate that of the latter, such that the optimized model has similar vulnerability to the target model \cite{iclr/Huang020,ijcai/XiaoLZHLS18}. The previous works which synthesize adversarial examples by the Normalizing Flow model are AdvFlow \cite{nips/DolatabadiEL20} and $\mathcal{CG}$-Attack \cite{cvpr/FengWFL0X22}. AdvFlow first map the input image to a hidden representation by the pre-trained Flow model and find a suitable disturbance in the hidden space, which then uses the natural evolution strategies (NES) to optimize the most helpful disturbance in an iterative updating manner. While $\mathcal{CG}$-Attack training a conditional Flow model (i.e., c-Glow \cite{aaai/LuH20}) relies on the local white box models with an additional adv loss first and then carries black-box attack with this well-trained flow model. Note that both AdvFlow and $\mathcal{CG}$-Attack are also query-based and need the models' whole outputs (soft-label) for attacking, which requires many queries or the more detailed outputs from the target model for performing a successful attack and is limited to attacking the physically deployed black-box models that only return the true label. 

The discussion above shows that the existing black-box attack methods mostly require thousands of queries and more detailed outputs on the target model to estimate the gradient and then carry out the attack iteratively to obtain a compelling adversarial example. In this situation, the attack is inefficient and impractical, while the time and computational consumption could be very considerable. In addition, the transferability of the adversarial examples obtained by querying and optimization is often limited; in other words, the generated adversarial examples are overly fit on the target model and are unqualified to attack other target models. When considering the attack under different datasets, the existing methods possess very limited capability to perform a successful attack. However, both the cross-model attack ability and the cross-dataset attack ability are sometimes valuable for real applications, especially when we do not have many chances to perform attack trials.

Therefore, the black-box attack poses the request for a method that is direct, efficient, and effective to perform attacks for different models and different datasets within limited queries and information. To achieve this goal, we know from the previous studies that the adversarial examples have a particular distribution related to the normal examples, and learning from such an adversarial distribution could help us to explore the vulnerability of different models. Hence, we are well motivated to develop a generative model that transfers from (or conditioned on) the distribution of clean examples to adversarial ones. It is also possible to achieve cross-dataset attacks by involving an increasing number of adversarial examples during offline learning.

\section{Preliminary}
\label{sec:preliminary}
Before introducing the details of the proposed framework, in this section, we first present the preliminary knowledge about adversarial attacks and normalizing flows.

\subsection{Adversarial Attack}
Given a well-trained DNN classifier $ f $ and a correctly classified input $ (\bm{x},y) \sim D $,  we have $ f(\bm{x})=y $, where $D$ denotes the accessible dataset. The adversarial example $ \bm{x}' $ is a neighbor of  $ \bm{x} $ and satisfies that $ f(\bm{x} ') \neq y$ and $ \left \| \bm{x}' - \bm{x} \right \|_p \leq \epsilon  $, where the $\ell_p$ norm is used as the metric function and $ \epsilon $ is usually a small value such as 8 and 16 with the image intensity [0, 255]. With this definition, the problem of finding an adversarial example becomes a constrained optimization problem:
\begin{equation}
      \label{eq:eq1}
      \bm{x}_{adv}= 
\underset{\left \| \bm{x}'-\bm{x} \right \|_p \leq \epsilon}{arg\ max\ \ell} (f(\bm{x}') \neq y),
\end{equation}
where $ \ell $ stands for a loss function that measures the confidence of the model outputs.

In the optimization-based methods, the above problem is solved by computing the gradients of the loss function in Eq. \ref{eq:eq1} to generate the adversarial example. By contrast, in this work, we formulate a statistical transformation from $P(\bm{x})$ to $P(\bm{x}')$ instead of involving an online optimization process.

\subsection{Normalizing Flow}
The normalizing flows \cite{corr/DinhKB14,nips/KingmaD18} are a class of probabilistic generative models, which are constructed based on a series of completely reversible components. The reversible property allows to transform from the original distribution to a new one and vice versa. By optimizing the model, a simple distribution (such as Gaussian distribution) can be transformed into a complex distribution of real data. The training process of normalizing flows is indeed an explicit likelihood maximization. Considering that the model is expressed by a fully invertible and differentiable function which transfers a random vector $ \bm{\bm{z}} $ from the Gaussian distribution to another vector $ \bm{x} $, we can employ such a model to generate high dimensional and complex data.

Specifically, given a reversible function $ {f:} \  \mathbb{R}^d\rightarrow \mathbb{R}^d $ and two random variables $ \bm{z}\sim p(\bm{z}) $ and $ \bm{z}'\sim p(\bm{z}') $ where $ \bm{z}' = f(\bm{z}) $, the change of variable rule tells that
\begin{equation}
      p(\bm{z}')=p(\bm{z})\left | det \frac{\partial {f^{-1}}}{\partial{\bm{z}'}} \right |,
\end{equation}
\begin{equation}
p(\bm{z})=p(\bm{z}')\left | det \frac{\partial {f}}{\partial{\bm{z}}} \right |,
\end{equation}
where $ det $ denotes the determinant operation. The above equation follows a chaining rule, in which a series of invertible mappings can be chained to approximate a sufficiently complex distribution, i.e.,
\begin{equation}
      \label{eq:eq2}
      \bm{z}_K = f_K \odot ... \odot  f_2 \odot f_1(\bm{z}_0),
\end{equation}
where each $ f $ is a reversible function called a flow step. Eq. \ref{eq:eq2} is the shorthand of $ f_K(f_{k-1}(...f_1(\bm{x}))) $. Assuming that $\bm{x}$ is the observed example and $\bm{z}$ is the hidden representation, we write the generative process as
\begin{equation}
      \bm{x}=f_{\theta}(\bm{z}),
\end{equation}
where $ f_{\theta} $ is the accumulate sum of all $ f $ in Eq. \ref{eq:eq2}. Based on the change-of-variables theorem, we write the log-density function of $ \bm{x}=\bm{z}_K $ as follows:
\begin{equation}
      \label{eq:eq21}
      -\log{p_K}(\bm{z}_K)=-\log p_0(\bm{z}_0)-\sum_{k=1}^{K}\log\left | det\frac{\partial \bm{z}_{k-1}}{\partial \bm{z}_{k}} \right |,
\end{equation}
where we use $\bm{z}_k=f_k(\bm{z}_{k-1})$ implicitly. The training process of normalizing flow is minimizing the above function, which exactly maximizes the likelihood on the observed training data. Hence, the optimization is stable and easy to implement. 

\subsection{Conditional Normalizing Flow}\label{cnf_intro}
In certain cases, the transformation between distributions is conditioned on external variables, for example a face is conditioned on age, gender, expression, etc. This has already been considered in the generative models such as CVAE \cite{nips/SohnLY15} and CGAN \cite{corr/MirzaO14}. In the flow-based models, the conditional normalizing flows allow us to involve the conditions in each flow step. Specifically, the reversible function $f$ accepts both the input variable $\bm{z}$ and the condition variable $c$ as inputs, which is formally expressed as $\bm{z}'=f(\bm{z};c)$, while the inverse mapping is $\bm{z}=f^{-1}(\bm{z}';c)$. By denoting the $K$-th flow step as $f_K$, the change of variables theorem says that
\begin{equation}
      \label{eq:eq9}
      \begin{split}
      &-\log{p_K}(\bm{z}_K;c)) \\ 
      &=-\log p_0(\bm{z}_0;c) -\sum_{k=1}^{K}\log \left | det\frac{\partial \bm{z}_{k-1}}{\partial f_k(\bm{z}_{k-1};c)} \right |.
      \end{split}
\end{equation}
Given a well-trained flow model, we first sample $\bm{z}_0$ from the Gaussian distribution and then perform a forward flow as
\begin{equation}
\bm{x}=f_\theta(\bm{z}_0;c).
\end{equation}
If we are interested in computing the probability density of an observed example $\bm{x}$, the inverse mapping is expressed as
\begin{equation}
\bm{z}_0=f^{-1}_\theta(\bm{x};c).
\end{equation}

\section{Methodology}
\label{sec:methodology}
In this section, we introduce the whole framework of the proposed adversarial attack in the generative manner, and the details of the model learning and inference.

\subsection{The DTA Framework}
Recall that the conventional attack methods generate the adversarial perturbation by performing a complex inference based on the target model, which is then added to the original example, resulting in the final adversarial example. This process is highly dependent on the inference result, which yields heavy computational cost and generally produces a single ``optimal'' example according to certain criteria. By contrast, in this paper, we start from a novel perspective and propose a novel generative adversarial attack method, which is called the distribution transform-based attack (DTA). Specifically, we advocate that all adversarial examples could follow a certain distribution that is misaligned with the normal distribution. This is mainly caused by the fixed training data involved in optimizing different deep models. In other words, the training data characterize a fixed distribution that is approximated by those models during training and hence, the distribution of the unseen data in training is common as well to the models. This explains why we consider that the adversarial examples (most of which are unseen data in training) follow a misaligned distribution. At this point, we reasonably assume a transformation from the distribution of normal examples to the distribution of adversarial examples. Since those two types of data exhibit similar appearances, the two distributions ideally overlap with each other and can be transformed mutually.

\begin{figure*}[ht]
    \setlength{\abovecaptionskip}{0.1cm}
    \setlength{\belowcaptionskip}{-10pt} 
      \centering
      \includegraphics[width=\textwidth]{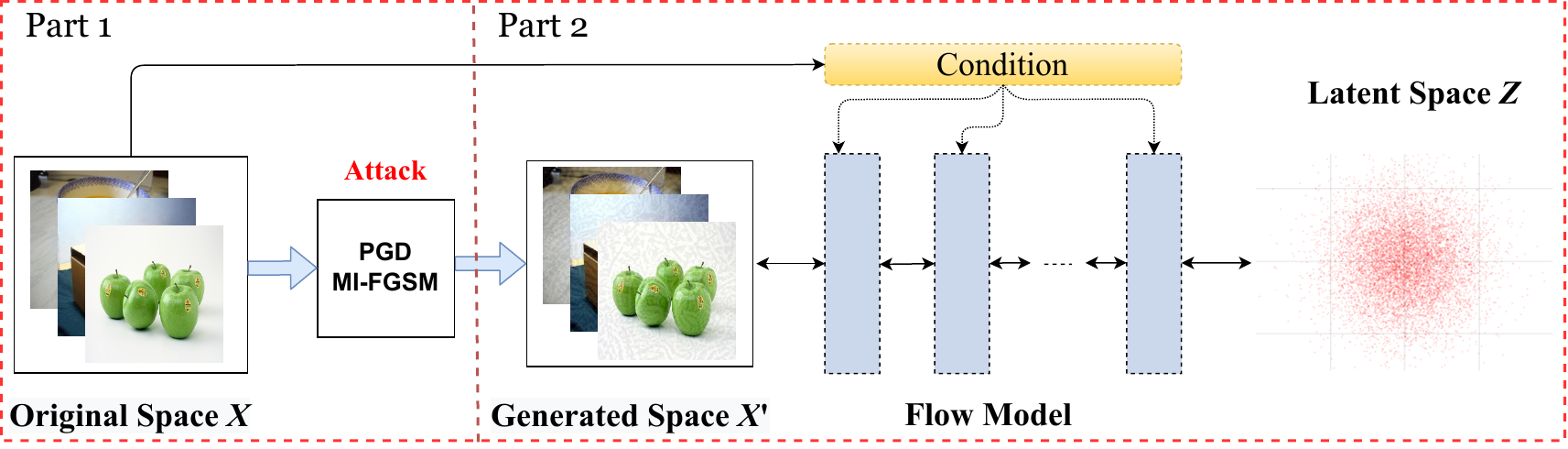}
      \caption{The framework of the Distribution transform-based attack. $ \bm{X}' $ is the adversarial space characterized by the collected adversarial examples and $ \bm{X} $ is the space of the corresponding original clean examples. The hidden space $ \bm{Z} $ follows a simple Gaussian distribution.}
      \label{fig:framwork}
\end{figure*}

The whole framework of the proposed method is illustrated in Fig. \ref{fig:framwork}. Based on the above discussion, we propose to collect a large number of adversarial examples $\bm{X}'$ by employing the existing white-box attack methods. While these examples look similar to the normal examples $\bm{X}$, a direct transformation between these two types of examples is nevertheless difficult or even prohibitive. This is because the small perturbation could be overwhelmed by the complex structures and textures in the normal example, and is therefore insensitive to the generation model. To alleviate this issue, we consider that the small perturbations should be conditioned on the normal inputs, which provide cues in the generative process. Specifically, the conditional normalizing flow is employed to implement the conditional generation process, which allows synthesizing the adversarial example based on the normal example and a random variable \cite{aaai/LuH20,cvpr/PumarolaPMF20,cvpr/LiuLGWL19}. The random variable could diversify the generated example, that is, when the flow model is well trained, we can randomly sample in the latent space $\bm{Z}$ to generate a batch of adversarial examples, which are inferenced forwardly by the flow model. The details of the flow model and the training and inference processes are discussed in the following sections.

\subsection{Conditional Normalizing Flow for Attack}
\begin{figure}[ht]
    \setlength{\abovecaptionskip}{0.1cm}
    \setlength{\belowcaptionskip}{-10pt} 
      \centering
    \label{fig:stru_DTA}
    \includegraphics[width=0.47\textwidth]{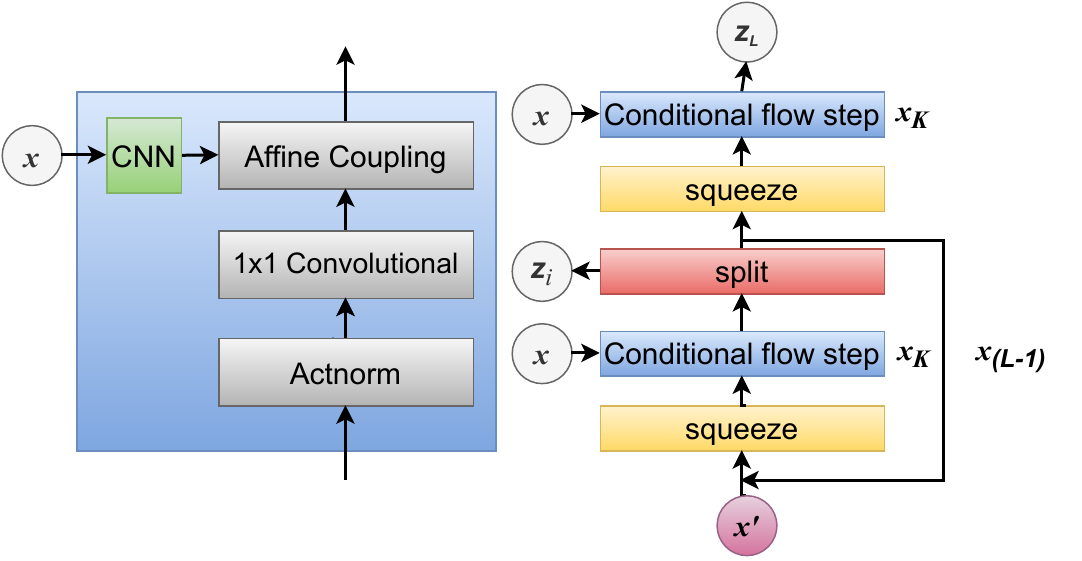}
      \caption{The conditional normalizing flow model for attack. The left depicts a basic conditional flow step, while the right plots the whole architecture.}
      \label{fig:stru}
\end{figure}

To implement a powerful normalizing flow that has a strong ability of processing image textures, we employ the basic GLOW model \cite{nips/KingmaD18}, which involves the convolutional operation, the coupling operation, and the normalization operation in the model construction. Since the original GLOW model does not consider conditions in the probabilistic modeling, we follow the work in \cite{corr/abs-1907-02392,aaai/LuH20} to properly integrate the image content in conditions. The architecture of the flow model is illustrated in Fig. \ref{fig:stru}. As seen, a basic flow step is a stack of the Actnorm layer, the 1x1 convolutional layer, and the affine coupling layer. A single flow block is constructed by cascading a squeeze layer, $K$ flow steps, and a split layer. Then, the whole architecture is built up by repeating the flow block for $L-1$ times, followed by the final layers, which consist of a squeeze layer and $K$ flow steps. The details of the Actnorm layer, the 1x1 convolutional layer, the affine coupling layer, and the squeeze and split layers can be found from GLOW \cite{nips/KingmaD18}. Regarding the conditions involved in each layer, it is proved that the original image is unsuitable to be directly fed to the condition. This is because the original image provides very low-level features which are insufficient for feature modeling and can burden the sub-networks in the affine coupling layer. Instead, high-level features are preferable. Hence, we follow the options in \cite{corr/abs-1907-02392,aaai/LuH20}, which suggests employing a pre-trained deep model to extract high-level features that are used as the condition. Specifically, we use the VGG-19 model pre-trained on CIFAR-10, SVHN, and ImageNet, respectively, and extract the features from the last conv layers. It is also possible to replace the VGG-19 model with other proper choices. During model training, the VGG-19 model can be fixed or optimized jointly with the flow model. In the current work, we fix this feature extraction model for simplicity.

\subsection{Adversarial Data Collection}
Recall that the adversarial examples obtained by using the existing white-box attack methods play a key role in the proposed framework. Hence, regarding how these examples are obtained, we present the details here instead of in the experiment section.

The concerned datasets in the current work include CIFAR-10 \cite{krizhevsky2009learning}, SVHN \cite{Netzer2011ReadingDI}, and ImageNet \cite{ijcv/RussakovskyDSKS15}, while the to-be-attacked models are also trained on these datasets. Specifically, the training sets of CIFAR-10 and SVHN are selected, while for ImageNet, we choose about 30,000 images from the validation set. All these data are used as normal examples by the white-box attack methods to generate adversarial examples. On CIFAR-10, the PGD method \cite{sp/Carlini017} is employed as the attacker, whereas the pre-trained ResNet-50 is employed as the target model. The MI-FGSM method based on multi-model integration \cite{cvpr/DongLPS0HL18} is employed on SVHN and ImageNet. For SVHN, ResNet-50 \cite{cvpr/HeZRS16}, InceptionV3 \cite{cvpr/SzegedyVISW16}, and SeNet-18 \cite{cvpr/HuSS18} are integrated, while the models are modified versions of the public ones\footnote{https://github.com/kuangliu/pytorch-cifar} and trained from scratch.
For ImageNet, InceptionV4 \cite{aaai/SzegedyIVA17}, InceptionResnetV2 \cite{aaai/SzegedyIVA17}, and ResNetV2-101 \cite{eccv/HeZRS16} are integrated, while the models are pre-trained and publicly accessible\footnote{https://github.com/tensorflow/models/tree/master/research/slim}. The adversarial examples are generated under two perturbation levels, including $ \epsilon = 8 $ and $ \epsilon = 16 $. The other hyperparameter settings of the attack methods follow the respective papers \cite{iclr/MadryMSTV18, icml/GuoGYWW19, nips/DolatabadiEL20}. In this way, we collect a batch of adversarial examples that will be used to optimize the proposed flow model. Note that the generated adversarial examples on a certain dataset are used to train the flow model that attacks the target model of the corresponding dataset. The cross-dataset attack is not applicable. 

To make a fair comparison in experiments, the normal examples for test are different from those for training. Specifically, for CIFAR-10 and SVHN, the test sets are employed as the input examples. For ImageNet, we randomly select 1000 images from the validation set, which are completely different from the 30,000 ones mentioned above.

\subsection{Training Details}
As introduced in Sec. \ref{cnf_intro}, the training of the conditional normalizing flow is to maximize the likelihood function on the training data with respect to the model parameters. Formally, assume that the collected adversarial example is denoted by $\bm{x}'\sim \bm{X}'$. The normal example is denoted by $\bm{x}\sim \bm{X}$, where the condition network produces the features as $c(\bm{x})$ (short for $c$). The hidden representation follows the Gaussian distribution, i.e. $\bm{z} \sim \mathcal{N}(0,1)$. The flow model is denoted by $f$, parameterized $\theta$, which have $\bm{x}'=f_\theta(\bm{z};c)$ and $\bm{z}=f^{-1}(\bm{x}';c)$. Then, the loss function to be minimized is expressed as
\begin{equation}
		\label{eq:mlloss}
		\begin{split}
		&	L(\theta;\bm{z},\bm{x}',c)=-\log p(\bm{x}'|\bm{z};c,\theta) \\			
		&	= -\log p_{\bm{z}}(f^{-1}_\theta(\bm{x}';c);c)-\log\left| det \frac{\partial f^{-1}_\theta(\bm{x}';c)}{\partial \bm{x}'}\right|,
		\end{split}
	\end{equation}
where the right-hand side of the above equation can be expanded layer-wisely according to Eq. \ref{eq:eq9}. By optimizing the above objective, the learned distribution $ p(\bm{x}'|\bm{z};c,\theta)$ characterizes the adversarial distribution as expected.

Considering that the interested task here is to generate an adversarial example that has a similar appearance to the example fed into the condition. Hence, we must ensure that the generation process from $\bm{z}$ to $\bm{x}'$ would bring no surprising result. To implement this, we impose an MSE loss in the training process. Specifically, the difference between the generated adversarial example $ \bm{x}' $ and the original input $ \bm{x}_{tr}^{'} $ is minimized according to
\begin{equation}
      \label{eq:eq10}
      L_{MSE}(\theta;\bm{z},c) = ||f_\theta(\bm{z};c) -  \bm{x}_{tr}^{'} ||_2,
\end{equation}
where $\bm{z}$ is randomly sampled from the Gaussian distribution in each training iteration. 

Note that the above losses in Eqs. \ref{eq:mlloss} and \ref{eq:eq10} consider the supervision in different spaces, where the former computes the loss in the hidden space while the latter concerns the adversarial space. Optimizing the losses simultaneously can bring unexpected effects since the loss propagation directions are conflicting. Hence, we propose to perform back-propagation based on the two losses alternatively. To be clear, in each iteration, we first update the model parameters based on Eq. \ref{eq:mlloss}. Then, given the input batch just used which contains $\bm{x}$, we randomly sample a batch of $\bm{z}$ and perform a forward flow to generate a batch of $ \bm{x}' $. The MSE loss between $ \bm{x}' $ and $  \bm{x}_{tr}^{'} $ is computed to update the model parameters, followed by the next iteration.

In the training process, we use the Adam algorithm to optimize the model parameters, while the learning rate is set as $10^{-4}$, the momentum is set to $ 0.999 $, and the maximal iteration number is 10,000.

\subsection{Generation of Adversarial Examples}
Given a well-trained flow model $f_\theta$, the hidden representations of the collected adversarial examples are expected to follow the assumed Gaussian distribution $\mathcal{N}(0, 1)$. But in practice, we find that these representations have shifted mean and standard deviation (std) values. This may be because the training data is insufficient. We may consider that the involved MSE loss could bias the center of the Gaussian distribution, but experiments tell that the shift occurs even without the MSE loss. Based on this observation, we also surprisingly find that sampling $\bm{z}$ based on the shifted mean and std values can bring improved performance than sampling from $\mathcal{N}(0, 1)$. Hence, before generating adversarial examples, we compute the hidden representations of all the training adversarial examples, which are used to calculate the mean value $\bm{\mu}$ and the std value $\sigma$, resulting in a new distribution $\mathcal{N}(\bm{\mu}, \sigma^2)$.

To generate an adversarial example, given an input normal example $\bm{x}$, we first randomly sample $\bm{z}$ from $\mathcal{N}(\bm{\mu}, \sigma^2)$ and then perform a forward process via $\bm{x}_{gen}=f_\theta(\bm{z}; c(\bm{x}))$. For the fairness of comparison, we follow the existing attack methods which constrain the perturbation within a certain range. Once we obtain the adversarial example $\bm{x}_{gen}$, we employ the clip function
\begin{equation}
\label{eq:clip}
\bm{x}'=Clip(\bm{x}+Clip( \bm{x}_{gen}-\bm{x}, -\epsilon, \epsilon),0,1)
\end{equation}
to ensure the imperceptible property of the perturbation, where $\epsilon$ is the acceptable noise budget during the attack. Two common cases are considered, including $ \epsilon =8 $ and $ \epsilon =16 $ for the pixel value $\in [0,255]$ (it will be scaled to $ \epsilon =8/255. $ and $ \epsilon =16/255. $ as the pixel value $\in [0,1]$ in code implementation).
\begin{algorithm}[htp]
    \caption{Distribution Transform-based Attack}
    \label{alg:alg1}
    \begin{algorithmic}[1]
        \Require
        $ \bm{X}_{tr} $:  a batch of clean examples used for training; $ \bm{X}'_{tr} $:  a batch of adversarial examples collected based on white-box attack methods, is used for training; $ \alpha $: the learning rate; $ T $: the maximal training iterations; $ Q $: the maximal querying number; $ \epsilon $: the noise budget; $ \bm{x}_{te} $:  a clean example used for test; $ M $: the target model to be attacked.
        
        \Ensure
        The adversarial example $\bm{x}'$ is used for attack.
        
        \Param
        The flow model $f_\theta$.
        
        \State Initialize the parameters of the flow model $f_\theta$;
        \For{$i=1$ to $T$}
        \State Optimize $f_\theta$ according to Eq. \ref{eq:mlloss};
        \State Optimize $f_\theta$ according to Eq. \ref{eq:eq10};
        \If{Convergence reached}
        \State break;
        \EndIf
        \EndFor
        \State Obtain optimized $f_\theta$;
        \State Compute the hidden representations of all examples in $\bm{X}'_{tr}$ via $ \bm{z}=f^{-1}(\bm{x}_{tr}^{'};c(\bm{x}_{tr}))$;
        \State Compute $ \bm{\mu} $ and $ \sigma$ from the hidden representations obtained in last step;
        \For {$i=1$ to $Q$}
        \State  Sample $\bm{z}$ from the distribution $ \mathcal N (\bm{\mu},\sigma^{2})$;
        \State Compute the adversarial example via  $\bm{x}’=f(\bm{z};c(\bm{x}_{te}))$;
        \State Clip the example via Eq. \ref{eq:clip};
        \If{Successfully attack $M$ by $\bm{x}'$}
        \State break.
        \EndIf
        \EndFor
        
    \end{algorithmic}

\end{algorithm}

The whole algorithm of DTA is listed in Alg. \ref{alg:alg1}, which could help readers to reimplement our method step-by-step.

\section{Experiments}
\label{Sec:experiments}
In this section, we evaluate the performance of the proposed DTA on black-box adversarial attacks through extensive experiments and comparisons. 



\begin{table*}
    \setlength{\abovecaptionskip}{0.1cm}
\caption{The performance comparison of black-box adversarial attack on the CIFAR-10 dataset, with the perturbation $\epsilon=8$ and $\epsilon=16$. We report the Attack Success Rate (ASR(\%)), Average Query Number, and Median Query Number under the max query limited in 100, 200, 300, 400 and 500, respectively.}
\let\cline\clineorig
\label{tab:cifar_asr}
\centering
\renewcommand{\arraystretch}{1.1}
\small
\resizebox*{1\linewidth}{!}{
\begin{tabular}{c|c|c|ccccc|ccccc|ccccc} 
\hline
\multirow{2}{*}{$\epsilon$} & \multirow{2}{*}{Target Model} & \multirow{2}{*}{Methods} & \multicolumn{5}{c|}{Attack Success Rate(\%)}                                       & \multicolumn{5}{c|}{Avg. Query Number}                                           & \multicolumn{5}{c}{Med. Query Number}                           \\
                            &                               &                          & 100            & 200            & 300            & 400            & 500            & 100           & 200           & 300            & 400            & 500            & 100        & 200        & 300        & 400        & 500         \\ 
\hline
\multirow{21}{*}{8}         & \multirow{7}{*}{VGG-16}       & Sign-OPT                 & -              & 0.54           & 3.54           & 2.56           & 3.74           & -             & 148.8         & 256.94         & 257.04         & 310.17         & -          & 144        & 264        & 255.5      & 264         \\
                            &                               & Bandits                  & 18.26          & 18.92          & 20.42          & 21.62          & 25.48          & 26.66         & 39.56         & 62.83          & 73.28          & 100.44         & 14         & 18         & 26         & 28         & 38          \\
                            &                               & Rays                     & 7.45           & 17.34          & 25.24          & 32.00          & 37.97          & 65.71         & 122.73        & 161.38         & 208.26         & 233.01         & 61.5       & 126        & 161        & 205        & 215         \\
                            &                               & Tangent & 3.04           & 3.62           & 3.91           & 4.44           & 4.95           & 21.01         & 48.84         & 48.84          & 87.31          & 141.95         & 28         & 28         & 28         & 28         & 28          \\
                            &                               & TA                       & 3.70           & 3.93           & 4.14           & 4.37           & 5.18           & 16.46         & 27.28         & 20.51          & 37.58          & 45.75          & 6          & 5          & 5          & 5          & 5           \\
                            &                               & CGBA                     & 2.11           & 3.33           & 5.49           & 6.62           & 7.64           & 71.91         & 99.68         & 153.87         & 190.72         & 230.89         & 89         & 93         & 149        & 159        & 216         \\
                            &                               & Ours                     & \textbf{69.35} & \textbf{71.23} & \textbf{72.36} & \textbf{73.00} & \textbf{73.57} & \textbf{5.76} & \textbf{9.45} & \textbf{12.40} & \textbf{15.02} & \textbf{19.02} & \textbf{3} & \textbf{5} & \textbf{4} & \textbf{5} & \textbf{5}  \\ 
\cline{2-18}
                            & \multirow{7}{*}{MobileNetv2}  & Sign-OPT                 & -              & 0.11           & 6.77           & 5.59           & 8.71           & -             & 152.00        & 251.86         & 256.60         & 295.20         & -          & 152        & 255        & 254        & 273         \\
                            &                               & Bandits                  & 36.53          & 45.32          & 47.38          & 51.62          & 51.85          & 24.84         & 43.54         & 66.49          & 51.62          & 103.84         & 14         & 18         & 30         & 40         & 50          \\
                            &                               & Rays                     & 12.11          & 30.38          & 43.72          & 56.84          & 59.91          & 68.48         & 123.19        & 168.73         & 207.71         & 235.22         & 66         & 124.5      & 170        & 208.5      & 224         \\
                            &                               & Tangent & 7.15           & 8.51           & 8.22           & 9.42           & 10.53          & 20.93         & 47.10         & 48.75         & 81.57          & 129.79         & 28         & 28         & 28         & 28         & 28          \\
                            &                               & TA                       & 8.24           & 8.42           & 8.54           & 8.87           & 8.97           & 12.48         & 21.30         & 21.46          & 22.98          & 27.62          & 5          & 5          & 5          & 5          & 5           \\
                            &                               & CGBA                     & 2.71           & 4.26           & 6.55           & 7.95           & 8.95           & 63.31         & 94.16         & 148.26         & 182.03         & 212.24         & 45         & 91         & 147        & 152        & 211         \\
                            &                               & Ours                     & \textbf{78.01} & \textbf{79.2}  & \textbf{80.48} & \textbf{81.31} & \textbf{81.53} & \textbf{4.45} & \textbf{6.91} & \textbf{9.93}  & \textbf{12.46} & \textbf{14.55} & \textbf{3} & \textbf{3} & \textbf{4} & \textbf{4} & \textbf{4}  \\ 
\cline{2-18}
                            & \multirow{7}{*}{ShuffleNetv2} & Sign-OPT                 & -              & 0.22           & 3.99           & 5.00           & 5.27           & -             & 186.00        & 260.03         & 265.8          & 288.69         & -          & 186        & 262        & 261        & 272         \\
                            &                               & Bandits                  & 26.91          & 36.80          & 39.80          & 43.97          & 49.72          & 30.66         & 48.70         & 75.78          & 103.57         & 113.88         & 20         & 26         & 46         & 52         & 58          \\
                            &                               & Rays                     & 12.16          & 26.93          & 37.19          & 48.57          & 55.6           & 67.88         & 121.65        & 153.27         & 197.38         & 224.49         & 66.5       & 124        & 148        & 190        & 194.00      \\
                            &                               & Tangent & 3.83           & 4.71           & 4.54           & 5.12           & 5.80           & 19.58         & 49.85         & 48.75          & 89.25          & 140.08         & 28         & 28         & 28         & 28         & 28          \\
                            &                               & TA                       & 5.42           & 6.42           & 6.69           & 7.34           & 8.25           & 17.98         & 15.98         & 39.62          & 41.53          & 46.82          & 5          & 5          & 6          & 6          & 5           \\
                            &                               & CGBA                     & 3.34           & 5.12           & 8.14           & 9.92           & 11.44          & 73.60         & 100.10        & 157.20         & 188.55         & 228.53         & 89         & 93         & 150        & 156        & 215         \\
                            &                               & Ours                     & \textbf{75.08} & \textbf{77.00} & \textbf{78.15} & \textbf{78.72} & \textbf{79.05} & \textbf{4.36} & \textbf{7.24} & \textbf{10.29} & \textbf{12.09} & \textbf{15.27} & \textbf{4} & \textbf{4} & \textbf{4} & \textbf{5} & \textbf{5}  \\ 
\hline
\multirow{21}{*}{16}        & \multirow{7}{*}{VGG-16}       & Sign-OPT                 & -              & 1.07           & 9.33           & 9.18           & 11.98          & -             & 151.60        & 259.14         & 267.55         & 286.43         & -          & 151        & 266        & 268.5      & 266.5       \\
                            &                               & Bandits                  & 54.14          & 58.86          & 66.27          & 63.96          & 66.52          & 27.14         & 39.71         & 45.45          & 56.68          & 63.47          & 18         & 24         & 22         & 24         & 28          \\
                            &                               & Rays                     & 21.17          & 43.12          & 60.66          & 70.90          & 80.32          & 65.70         & 118.04        & 161.62         & 187.88         & 211.81         & 65         & 118.5      & 162        & 184        & 196         \\
                            &                               & Tangent & 14.72          & 18.45          & 18.04          & 20.92          & 22.92          & 22.05         & 50.09         & 50.93          & 92.62          & 134.20         & 28         & 28         & 28         & 28         & 28          \\
                            &                               & TA                       & 16.63          & 17.26          & 18.51          & 20.27          & 22.74          & 14.84         & 18.32         & 19.27          & 27.90          & 33.79          & 5          & 5          & 5          & 5          & 5           \\
                            &                               & CGBA                     & 4.10           & 6.67           & 10.07          & 12.21          & 13.85          & 58.28         & 90.84         & 145.05         & 183.00         & 216.49         & 41         & 89         & 144        & 151        & 211         \\
                            &                               & Ours                     & \textbf{75.32} & \textbf{77.95} & \textbf{79.11} & \textbf{79.96} & \textbf{80.59} & \textbf{4.53} & \textbf{7.93} & \textbf{11.86} & \textbf{13.43} & \textbf{15.43} & \textbf{1} & \textbf{1} & \textbf{1} & \textbf{1} & \textbf{1}  \\ 
\cline{2-18}
                            & \multirow{7}{*}{MobileNetv2}  & Sign-OPT                 & -              & 1.19           & 16.99          & 20.44          & 25.59          & -             & 178.09        & 250.44         & 262.80         & 298.54         & -          & 174        & 255        & 264        & 268         \\
                            &                               & Bandits                  & 74.02          & 79.16          & 85.55          & 84.58          & 87.10          & 23.04         & 29.08         & 34.68          & 43.22          & 48.94          & 16         & 16         & 16         & 18         & 16          \\
                            &                               & Rays                     & 34.64          & 60.60          & 74.87          & 85.35          & 89.71          & 69.67         & 108.94        & 133.98         & 152.75         & 179.82         & 69         & 104        & 117        & 132        & 149         \\
                            &                               & Tangent        & 27.57          & 32.84          & 33.29          & 36.68          & 39.01          & 21.36         & 47.12         & 50.15          & 77.49          & 126.86         & 28         & 28         & 28         & 28         & 28          \\
                            &                               & TA                       & 31.92          & 33.44          & 38.63          & 39.45          & 42.41          & 12.60         & 18.01         & 22.70          & 27.31          & 32.31          & 5          & 5          & 5          & 5          & 5           \\
                            &                               & CGBA                     & 7.59           & 10.06          & 14.88          & 16.69          & 19.43          & 53.71         & 77.82         & 131.22         & 158.47         & 200.29         & 40         & 44.5       & 97         & 143        & 158         \\
                            &                               & Ours                     & \textbf{86.14} & \textbf{88.7}  & \textbf{89.66} & \textbf{90.30} & \textbf{91.05} & \textbf{4.26} & \textbf{7.78} & \textbf{9.79}  & \textbf{11.93} & \textbf{13.90} & \textbf{1} & \textbf{1} & \textbf{1} & \textbf{1} & \textbf{1}  \\ 
\cline{2-18}
                            & \multirow{7}{*}{ShuffleNetv2} & Sign-OPT                 & -              & 0.11           & 9.70           & 11.62          & 15.26          & -             & 183.00        & 261.43         & 268.28         & 309.84         & -          & 183        & 263        & 266        & 275         \\
                            &                               & Bandits                  & 63.19          & 73.59          & 74.76          & 76.68          & 77.23          & 26.82         & 38.57         & 48.33          & 57.23          & 63.78          & 20         & 20         & 24         & 24         & 24          \\
                            &                               & Rays                     & 31.79          & 58.19          & 74.73          & 83.37          & 86.11          & 63.64         & 108.71        & 138.37         & 156.16         & 179.90         & 63         & 102        & 129        & 139        & 146.5       \\
                            &                               & Tangent           & 16.28          & 20.20          & 20.00          & 22.68          & 25.17          & 21.59         & 49.42         & 52.29          & 90.03          & 131.56         & 28         & 28         & 28         & 28         & 28          \\
                            &                               & TA                       & 23.77          & 25.80          & 26.79          & 28.80          & 29.48          & 15.00         & 25.43         & 30.14          & 36.07          & 42.08          & 5          & 5          & 5.5        & 5          & 5           \\
                            &                               & CGBA                     & 6.37           & 10.05          & 15.29          & 18.12          & 20.81          & 59.07         & 87.28         & 148.99         & 182.12         & 218.26         & 42         & 87         & 147        & 152        & 212         \\
                            &                               & Ours                     & \textbf{82.11} & \textbf{84.91} & \textbf{86.1}  & \textbf{87.29} & \textbf{87.82} & \textbf{5.28} & \textbf{8.60} & \textbf{12.02} & \textbf{15.07} & \textbf{17.05} & \textbf{1} & \textbf{1} & \textbf{1} & \textbf{1} & \textbf{1}  \\
\hline
\end{tabular}
}
\end{table*}

\begin{table*}[htp]
\caption{The performance comparison of black-box adversarial attack on the SVHN dataset, with the perturbation $\epsilon=8$ and $\epsilon=16$. We report the Attack Success Rate (ASR(\%)), Average Query Number, and Median Query Number under the max query limited in 100, 200, 300, 400 and 500, respectively.}
\let\cline\clineorig
	\label{tab:svhn_asr}
\centering
\renewcommand{\arraystretch}{1.1}
\small
\resizebox*{1\linewidth}{!}{
\begin{tabular}{c|c|c|ccccc|ccccc|ccccc} 
\hline
\multirow{2}{*}{$\epsilon$} & \multirow{2}{*}{Target Model} & \multirow{2}{*}{Methods} & \multicolumn{5}{c|}{Attack Success Rate(\%)}                                       & \multicolumn{5}{c|}{Avg. Query Number}                                             & \multicolumn{5}{c}{Med. Query Number}                           \\
                            &                               &                          & 100            & 200            & 300            & 400            & 500            & 100            & 200            & 300            & 400            & 500            & 100        & 200        & 300        & 400        & 500         \\ 
\hline
\multirow{21}{*}{8}         & \multirow{7}{*}{VGG-16}       & Sign-OPT                 & -             & -             & 1.85           & 2.27           & 1.30           & -             & -             & 257.24         & 261.00         & 325.33         & -         & -         & 257        & 253        & 288         \\
                            &                               & Bandits                  & 11.22          & 20.71          & 24.23          & 26.85          & 32.19          & 33.79          & 60.30          & 103.77         & 13.61          & 145.20         & 26         & 40         & 76         & 106        & 104         \\
                            &                               & Rays                     & 9.77           & 22.72          & 35.87          & 45.84          & 51.50          & 69.34          & 126.15         & 172.37         & 207.28         & 234.90         & 76.5       & 131        & 175        & 205        & 221         \\
                            &                               & Tangent           & 1.35           & 1.51           & 1.56           & 1.74           & 1.88           & 20.28          & 50.47          & 45.73          & 88.24          & 139.75         & 28         & 28         & 28         & 28         & 28          \\
                            &                               & TA                       & 4.30           & 4.41           & 4.63           & 4.77           & 5.05           & 29.93          & 42.65          & 48.88          & 53.91          & 57.58          & 19         & 33.5       & 18         & 21         & 20.5        \\
                            &                               & CGBA                     & 5.21           & 9.60           & 13.81          & 18.22          & 19.09          & 78.42          & 107.65         & 155.97         & 189.58         & 215.50         & 94         & 99         & 154        & 161        & 217         \\
                            &                               & Ours                     & \textbf{47.62} & \textbf{51.29} & \textbf{53.03} & \textbf{54.44} & \textbf{55.36} & \textbf{6.11}  & \textbf{10.96} & \textbf{15.86} & \textbf{20.45} & \textbf{24.66} & \textbf{4} & \textbf{5} & \textbf{5} & \textbf{6} & \textbf{7}  \\ 
\cline{2-18}
                            & \multirow{7}{*}{MobileNetv2}  & Sign-OPT                 & -             & -             & 2.42           & 1.86           & 1.44           & -              & -             & 256.52         & 257.56         & 290.21         & -         & -         & 250        & 257        & 260         \\
                            &                               & Bandits                  & 10.90          & 14.71          & 21.11          & 25.29          & 26.26          & 35.77          & 72.76          & 114.28         & 139.97         & 173.94         & 32         & 52         & 96         & 101        & 138         \\
                            &                               & Rays                     & 5.22           & 16.75          & 27.33          & 36.07          & 45.31          & 71.50          & 138.94         & 179.41         & 225.09         & 255.95         & 71         & 140.5      & 181        & 232        & 249         \\
                            &                               & Tangent           & 0.98           & 1.11           & 1.14           & 1.27           & 1.32           & 20.07          & 48.55          & 46.87          & 88.34          & 116.83         & 28         & 28         & 28         & 28         & 28          \\
                            &                               & TA                       & 2.19           & 2.11           & 2.77           & 2.69           & 3.01           & 19.38          & 40.50          & 37.82          & 45.42          & 45.07          & 5          & 25.5       & 16         & 16         & 7           \\
                            &                               & CGBA                     & 3.94           & 7.38           & 11.04          & 13.08          & 14.92          & 78.50          & 109.73         & 160.55         & 188.97         & 217.22         & 94.5       & 99         & 155        & 161        & 218         \\
                            &                               & Ours                     & \textbf{37.66} & \textbf{41.09} & \textbf{43.10} & \textbf{44.53} & \textbf{47.47} & \textbf{6.06}  & \textbf{11.29} & \textbf{15.77} & \textbf{20.72} & \textbf{25.80} & \textbf{5} & \textbf{6} & \textbf{7} & \textbf{7} & \textbf{7}  \\ 
\cline{2-18}
                            & \multirow{7}{*}{ShuffleNetv2} & Sign-OPT                 & -             & -              & 1.89           & 2.02           & 1.89           & -             & -              & 254.06         & 248.90         & 268.45         & -         & -          & 257        & 247        & 254         \\
                            &                               & Bandits                  & 9.35           & 14.93          & 20.84          & 20.75          & 27.42          & 38.09          & 72.95          & 110.80         & 130.09         & 172.42         & 30         & 60         & 80         & 110        & 122         \\
                            &                               & Rays                     & 7.32           & 18.49          & 35.10          & 41.40          & 47.08          & 72.32          & 131.56         & 175.13         & 227.31         & 255.19         & 74         & 137        & 176        & 234        & 244.5       \\
                            &                               & Tangent           & 0.93           & 0.98           & 1.13           & 1.18           & 1.27           & 18.82          & 50.48          & 43.97          & 89.16          & 103.99         & 11         & 28         & 28         & 28         & 28          \\
                            &                               & TA                       & 2.62           & 2.20           & 2.21           & 2.49           & 2.56           & 21.36          & 24.48          & 43.81          & 62.21          & 46.00          & 7          & 6          & 23         & 32.5       & 16          \\
                            &                               & CGBA                     & 3.97           & 7.59           & 12.25          & 14.92          & 16.58          & 78.79          & 107.34         & 158.09         & 192.36         & 216.83         & 95         & 100        & 156        & 161        & 218         \\
                            &                               & Ours                     & \textbf{40.09} & \textbf{43.76} & \textbf{45.63} & \textbf{46.78} & \textbf{49.76} & \textbf{5.97}  & \textbf{11.05} & \textbf{15.82} & \textbf{20.07} & \textbf{24.70} & \textbf{5} & \textbf{7} & \textbf{8} & \textbf{8} & \textbf{9}  \\ 
\hline
\multirow{21}{*}{16}        & \multirow{7}{*}{VGG-16}       & Sign-OPT                 & -             & -             & 3.81           & 5.93           & 5.63           & -             & -             & 257.51         & 259.55         & 297.23         & -         & -         & 257        & 261        & 263         \\
                            &                               & Bandits                  & 28.14          & 33.51          & 34.05          & 41.51          & 42.11          & 32.51          & 50.42          & 65.74          & 104.86         & 130.52         & 24         & 32         & 36         & 50         & 62          \\
                            &                               & Rays                     & 23.98          & 53.68          & 70.91          & 81.58          & \textbf{85.13}          & 70.65          & 116.16         & 152.35         & 173.04         & 194.00         & 68.5       & 115        & 148        & 165        & 173         \\
                            &                               & Tangent           & 4.24           & 5.32           & 5.03           & 6.04           & 6.61           & 20.57          & 52.42          & 48.75          & 97.55          & 148.16         & 28         & 28         & 28         & 28         & 28          \\
                            &                               & TA                       & 10.65          & 12.35          & 13.16          & 14.39          & 15.33          & 25.55          & 34.89          & 53.00          & 61.44          & 55.61          & 12         & 15         & 25.5       & 25.5       & 14          \\
                            &                               & CGBA                     & 9.29           & 17.48          & 26.95          & 31.91          & 36.33          & 77.52          & 108.66         & 160.72         & 194.79         & 227.29         & 93         & 99         & 155        & 161        & 220         \\
                            &                               & Ours                     & \textbf{71.61} & \textbf{72.63} & \textbf{74.39} & \textbf{78.41} & 84.11 & \textbf{10.04} & \textbf{15.87} & \textbf{22.28} & \textbf{27.64} & \textbf{37.28} & \textbf{3} & \textbf{3} & \textbf{4} & \textbf{4} & \textbf{4}  \\ 
\cline{2-18}
                            & \multirow{7}{*}{MobileNetv2}  & Sign-OPT                 & -             & -             & 4.64           & 4.44           & 4.32           & -             & -             & 261.43         & 260.54         & 278.17         & -         & -         & 262        & 260        & 258         \\
                            &                               & Bandits                  & 22.75          & 29.23          & 35.77          & 35.18          & 40.33          & 37.86          & 60.04          & 76.75          & 101.23         & 126.68         & 32         & 36         & 44         & 53         & 66          \\
                            &                               & Rays                     & 13.92          & 33.65          & 55.86          & 68.38          & 75.63          & 73.39          & 125.95         & 170.54         & 205.98         & 233.09         & 72         & 131        & 166        & 201.5      & 210.5       \\
                            &                               & Tangent           & 3.40           & 3.68           & 3.96           & 4.24           & 4.73           & 21.15          & 48.75          & 54.76          & 94.45          & 133.37         & 28         & 28         & 28         & 28         & 28          \\
                            &                               & TA                       & 5.38           & 7.41           & 6.15           & 7.32           & 8.12           & 22.49          & 28.70          & 35.22          & 45.56          & 63.19          & 8          & 8          & 13         & 11         & 16          \\
                            &                               & CGBA                     & 7.47           & 13.75          & 22.74          & 28.06          & 31.83          & 80.14          & 110.20         & 168.95         & 203.78         & 232.06         & 94         & 100        & 157        & 216        & 221         \\
                            &                               & Ours                     & \textbf{65.41} & \textbf{68.07} & \textbf{71.90} & \textbf{73.27} & \textbf{78.18} & \textbf{12.64} & \textbf{17.55} & \textbf{21.30} & \textbf{27.79} & \textbf{30.24} & \textbf{4} & \textbf{4} & \textbf{3} & \textbf{3} & \textbf{4}  \\ 
\cline{2-18}
                            & \multirow{7}{*}{ShuffleNetv2} & Sign-OPT                 & -             & 0.11           & 4.41           & 4.57           & 5.24           & -             & 197.00         & 255.91         & 259.00         & 308.15         & -         & 197        & 255.5      & 254        & 276.5       \\
                            &                               & Bandits                  & 22.97          & 28.07          & 32.26          & 35.24          & 38.21          & 36.35          & 64.62          & 84.43          & 104.09         & 125.30         & 30         & 44         & 48         & 54         & 62          \\
                            &                               & Rays                     & 17.22          & 44.02          & 64.49          & 74.63          & \textbf{80.19} & 74.07          & 125.03         & 174.63         & 193.11         & 222.37         & 76         & 127        & 176        & 180.5      & 208         \\
                            &                               & Tangent           & 2.81           & 3.15           & 3.20           & 3.91           & 4.19           & 20.93          & 53.41          & 51.26          & 88.87          & 149.96         & 28         & 28         & 28         & 28         & 29          \\
                            &                               & TA                       & 5.68           & 7.14           & 8.33           & 7.45           & 7.79           & 29.20          & 38.74          & 45.35          & 47.85          & 73.89          & 16.5       & 17.5       & 17.5       & 22         & 32          \\
                            &                               & CGBA                     & 7.71           & 14.59          & 23.36          & 29.14          & 32.75          & 78.06          & 107.36         & 160.92         & 203.51         & 234.70         & 94         & 99         & 156        & 215        & 222         \\
                            &                               & Ours                     & \textbf{67.99} & \textbf{72.84} & \textbf{75.74} & \textbf{77.26} & 78.72          & \textbf{10.68} & \textbf{17.57} & \textbf{19.83} & \textbf{26.07} & \textbf{38.08} & \textbf{4} & \textbf{5} & \textbf{4} & \textbf{4} & \textbf{4}  \\
\hline
\end{tabular}
}
\end{table*}

\subsection{Settings}
As mentioned previously, three popular datasets are considered, including CIFAR-10 \cite{cifar-10}, SVHN \cite{2011Reading}, and ImageNet \cite{ijcv/RussakovskyDSKS15}. 

Regarding the target models to be attacked, we employ the public models pre-trained on the corresponding datasets or the models that are trained from scratch if not publicly accessible. Specifically, we mainly target models include the VGG-16 \cite{corr/SimonyanZ14a}, the MobileNetV2 \cite{corr/abs-1801-04381}, and the ShuffleNetV2 \cite{eccv/MaZZS18}. For CIFAR-10 and ImageNet, we use their pre-trained weights from the GitHub repository \textit{pytorch-cifar-models}\footnote{https://github.com/chenyaofo/pytorch-cifar-models} and the PyTorch \footnote{https://github.com/pytorch}, respectively. While for SVHN, we trained these models from scratch, where the training process of each model is stopped until the best performance is obtained, in which condition the classification accuracy on the test set is above 90\%. 

To objectively evaluate the performance of the proposed framework, we make a comparison with the related state-of-the-art decision-based (hard-label) methods, including Bandits \cite{iclr/IlyasEM19}, Sign-OPT \cite{iclr/ChengSCC0H20}, Rays \cite{kdd/ChenG20}, Tangent Attack (Tangent) \cite{nips/MaGCYW21}, Triangle Attack (TA) \cite{eccv/WangZTGHLL22} and CGBA \cite{Reza_2023_ICCV}. The implementations of these methods are based on the released codes with default settings in the corresponding papers. The proposed DTA is implemented by using the PyTorch framework. To make a quantitative comparison, we use the metrics of attack success rate (ASR), average query count and median query count as the previous works use \cite{kdd/ChenG20, pami/DongCPSZ22}.

All the experiments are conducted on a GPU server with a single Tesla V100 32GB GPU, 2 x Xeon Silver 4208 CPU, and RAM 256GB.

\begin{table*}
\centering
\caption{The performance comparison of black-box adversarial attack on the ImageNet dataset, with the perturbation $\epsilon=8$ and $\epsilon=16$. We report the Attack Success Rate (ASR(\%)), Average Query Number, and Median Query Number under the max query limited in 100, 200, 300, 400 and 500, respectively.}
\let\cline\clineorig
\label{tab:imagenet}
\small
\renewcommand{\arraystretch}{1.1}
\setlength\tabcolsep{4pt}
\resizebox*{1\linewidth}{!}{
\begin{tabular}{c|c|c|ccccc|ccccc|ccccc} 
\hline
\multirow{2}{*}{$\epsilon$} & \multirow{2}{*}{Target Model} & \multirow{2}{*}{Methods} & \multicolumn{5}{c|}{Attack Success Rate(\%)}                                       & \multicolumn{5}{c|}{Avg. Query Number}                                             & \multicolumn{5}{c}{Med. Query Number}                           \\
                            &                               &                          & 100            & 200            & 300            & 400            & 500            & 100            & 200            & 300            & 400            & 500            & 100        & 200        & 300        & 400        & 500         \\ 
\hline
\multirow{21}{*}{8}         & \multirow{7}{*}{VGG-16}       & Sign-OPT                 & -             & -             & 4.77           & 4.49           & 4.77           & -             & -             & 261.09         & 273.72         & 306.38         & -         & -         & 260.5      & 264        & 272.5       \\
                            &                               & Bandits                  & 21.35          & 26.63          & 30.28          & 45.84          & 48.28          & 26.92          & 52.78          & 87.87          & 122.02         & 140.74         & 12         & 26         & 56         & 75         & 84          \\
                            &                               & Rays                     & 13.18          & 24.72          & 35.53          & 45.58          & 48.25          & 70.41          & 114.20         & 159.00         & 202.66         & 235.97         & 72         & 110        & 152        & 199        & 229         \\
                            &                               & Tangent           & 4.63           & 4.91           & 5.47           & 4.77           & 5.61           & 19.56          & 39.27          & 34.04          & 69.47          & 27.10          & 11         & 11         & 11         & 11         & 11          \\
                            &                               & TA                       & 23.70          & 26.23          & 27.35          & 28.61          & 28.47          & 21.04          & 31.01          & 40.86          & 56.43          & 53.12          & 11         & 16         & 18         & 18         & 18          \\
                            &                               & CGBA                     & 5.89           & 9.68           & 12.90          & 18.23          & 22.16          & 75.98          & 103.77         & 145.72         & 205.66         & 246.06         & 93.5       & 98         & 155        & 191.5      & 232         \\
                            &                               & Ours                     & \textbf{24.86} & \textbf{28.37} & \textbf{33.06} & \textbf{39.79} & \textbf{44.21} & \textbf{9.98}  & \textbf{28.01} & \textbf{25.06} & \textbf{43.20} & \textbf{44.46} & \textbf{2} & \textbf{3} & \textbf{3} & \textbf{4} & \textbf{4}  \\ 
\cline{2-18}
                            & \multirow{7}{*}{MobileNetv2}  & Sign-OPT                 & -             & -             & 6.34           & 6.90           & 7.61           & -             & -             & 265.78         & 269.82         & 279.50         & -         & -         & 264        & 269        & 267.5       \\
                            &                               & Bandits                  & 28.17          & 28.38          & 36.49          & 33.85          & 39.68          & 23.83          & 51.22          & 85.94          & 112.80         & 131.95         & 6          & 17         & 42         & 64         & 67          \\
                            &                               & Rays                     & 20.14          & 29.44          & 33.86          & 45.15          & 61.35          & 65.73          & 95.56          & 150.64         & 194.50         & 220.72         & 60         & 83         & 134        & 196        & 224         \\
                            &                               & Tangent           & 5.92           & 6.20           & 7.32           & 5.49           & 6.76           & 14.35          & 23.00          & 18.22          & 26.93          & 42.14          & 11         & 11         & 11         & 11         & 11          \\
                            &                               & TA                       & 22.54          & 24.37          & 24.93          & 26.48          & 27.46          & 20.12          & 37.42          & 42.45          & 49.54          & 65.22          & 10         & 12         & 11         & 14         & 18          \\
                            &                               & CGBA                     & 5.21           & 8.87           & 12.54          & 17.32          & 19.15          & 59.16          & 91.32          & 138.74         & 192.17         & 215.53         & 45         & 95         & 104        & 162        & 165         \\
                            &                               & Ours                     & \textbf{29.77} & \textbf{35.64} & \textbf{38.94} & \textbf{47.71} & \textbf{49.82} & \textbf{10.08} & \textbf{24.51} & \textbf{35.06} & \textbf{31.71} & \textbf{40.36} & \textbf{2} & \textbf{3} & \textbf{4} & \textbf{3} & \textbf{3}  \\ 
\cline{2-18}
                            & \multirow{7}{*}{ShuffleNetv2} & Sign-OPT                 & -             & -             & 7.03           & 8.17           & 9.80           & -             & -             & 264.16         & 266.86         & 313.85         & -         & -         & 262        & 262        & 273         \\
                            &                               & Bandits                  & 46.08          & 48.63          & 53.46          & 57.74          & 61.54          & 25.14          & 46.79          & 64.81          & 80.80          & 105.21         & 8          & 18         & 26         & 29         & 40          \\
                            &                               & Rays                     & 30.61          & 47.71          & 58.27          & 60.98          & \textbf{69.61}          & 64.88          & 100.25         & 129.83         & 152.97         & 173.68         & 60         & 90.5       & 109        & 127        & 141         \\
                            &                               & Tangent           & 8.01           & 8.66           & 8.01           & 8.33           & 8.17           & 12.69          & 24.27          & 27.62          & 28.17          & 57.64          & 11         & 11         & 11         & 11         & 11          \\
                            &                               & TA                       & 32.35          & 36.11          & 38.56          & 40.36          & 41.99          & 22.46          & 31.78          & 41.78          & 47.26          & 56.74          & 10.05      & 10         & 15         & 19         & 18          \\
                            &                               & CGBA                     & 7.84           & 13.56          & 18.14          & 23.53          & 27.94          & 72.75          & 117.12         & 155.18         & 194.99         & 241.50         & 93         & 106        & 157        & 160        & 228         \\
                            &                               & Ours                     & \textbf{50.82} & \textbf{53.76} & \textbf{58.74} & \textbf{61.21} & 65.88 & \textbf{7.73}  & \textbf{12.11} & \textbf{17.44} & \textbf{19.97} & \textbf{31.31} & \textbf{1} & \textbf{2} & \textbf{2} & \textbf{2} & \textbf{2}  \\ 
\hline
\multirow{21}{*}{16}        & \multirow{7}{*}{VGG-16}       & Sign-OPT                 & -             & -             & 7.43           & 9.40           & 10.94          & -             & -             & 258.72         & 274.02         & 289.83         & -         & -         & 257        & 270        & 269         \\
                            &                               & Bandits                  & 47.14          & 63.64          & 60.94          & 58.11          & 76.06          & 26.36          & 46.69          & 66.86          & 83.26          & 107.32         & 14         & 24         & 30         & 32         & 44          \\
                            &                               & Rays                     & 29.92          & 45.14          & 61.34          & 69.61          & 74.16          & 69.80          & 106.72         & 144.76         & 168.45         & 195.82         & 72         & 101        & 131        & 148.5      & 163         \\
                            &                               & Tangent           & 11.78          & 11.78          & 11.64          & 12.90          & 13.60          & 17.76          & 31.93          & 36.09          & 52.86          & 57.12          & 11         & 11         & 11         & 11         & 11          \\
                            &                               & TA                       & 53.72          & 60.45          & 61.29          & 59.19          & 61.99          & 15.12          & 25.61          & 36.08          & 36.19          & 42.51          & 8          & 10         & 13         & 10         & 10          \\
                            &                               & CGBA                     & 10.80          & 16.83          & 24.26          & 31.98          & 36.47          & 66.95          & 95.88          & 141.29         & 194.05         & 227.48         & 54         & 96         & 151        & 163        & 224         \\
                            &                               & Ours                     & \textbf{64.97} & \textbf{67.44} & \textbf{70.79} & \textbf{71.20} & \textbf{78.84} & \textbf{9.05}  & \textbf{11.95} & \textbf{18.09} & \textbf{17.83} & \textbf{28.58} & \textbf{1} & \textbf{1} & \textbf{1} & \textbf{1} & \textbf{1}  \\ 
\cline{2-18}
                            & \multirow{7}{*}{MobileNetv2}  & Sign-OPT                 & -             & -             & 10.56          & 11.13          & 13.52          & -             & -             & 264.23         & 272.06         & 294.81         & -         & -         & 263        & 268        & 269         \\
                            &                               & Bandits                  & 34.85          & 42.42          & 52.31          & 56.84          & 68.12          & 23.65          & 47.20          & 66.63          & 87.53          & 115.71         & 10         & 14         & 20         & 27         & 42          \\
                            &                               & Rays                     & 20.14          & 29.44          & 43.86          & 55.15          & 61.35          & 65.73          & 95.56          & 150.64         & 194.50         & 220.72         & 60         & 93         & 134        & 196        & 224         \\
                            &                               & Tangent           & 11.69          & 12.25          & 12.25          & 12.96          & 14.08          & 15.97          & 21.68          & 39.32          & 42.07          & 77.54          & 11         & 11         & 11         & 11         & 11          \\
                            &                               & TA                       & 46.48          & 48.87          & 50.85          & 51.13          & 52.39          & 20.66          & 29.80          & 36.10          & 40.07          & 51.54          & 7          & 11         & 11         & 12         & 14          \\
                            &                               & CGBA                     & 10.99          & 15.77          & 20.85          & 26.34          & 31.97          & 55.63          & 82.38          & 113.97         & 167.17         & 209.34         & 45         & 52.5       & 95.5       & 153        & 161         \\
                            &                               & Ours                     & \textbf{54.81} & \textbf{58.02} & \textbf{59.39} & \textbf{60.92} & \textbf{63.74} & \textbf{9.75}  & \textbf{15.63} & \textbf{23.40} & \textbf{24.62} & \textbf{26.46} & \textbf{1} & \textbf{1} & \textbf{1} & \textbf{1} & \textbf{1}  \\ 
\cline{2-18}
                            & \multirow{7}{*}{ShuffleNetv2} & Sign-OPT                 & -             & -              & 15.69          & 17.49          & 19.94          & -             & -              & 266.94         & 270.48         & 306.49         & -         & -          & 268        & 267        & 277.5       \\
                            &                               & Bandits                  & 43.33          & 46.65          & 55.85          & 64.70          & \textbf{72.83}          & 19.71          & 33.76          & 44.29          & 55.39          & 64.94          & 10         & 12         & 12         & 14         & 16          \\
                            &                               & Rays                     & 30.08          & 37.69          & 43.01          & 48.40          & 61.50          & 59.76          & 89.20          & 112.59         & 127.91         & 139.31         & 57         & 75         & 84         & 96         & 102.5       \\
                            &                               & Tangent           & 23.20          & 22.39          & 22.88          & 23.69          & 23.37          & 17.01          & 33.34          & 32.24          & 50.87          & 52.71          & 11         & 11         & 11         & 11         & 11          \\
                            &                               & TA                       & 43.73          & 46.99          & 50.59          & 53.26          & 64.02          & 15.23          & 22.92          & 31.56          & 29.37          & 42.94          & 5          & 6          & 6          & 6          & 8           \\
                            &                               & CGBA                     & 14.38          & 22.39          & 32.35          & 37.58          & 42.81          & 71.03          & 94.04          & 13.76          & 180.45         & 214.59         & 91.5       & 95         & 131        & 157        & 218         \\
                            &                               & Ours                     & \textbf{47.31} & \textbf{54.85} & \textbf{58.21} & \textbf{66.54} & 71.67 & \textbf{8.18}  & \textbf{12.51} & \textbf{22.28} & \textbf{26.80} & \textbf{33.01} & \textbf{1} & \textbf{1} & \textbf{1} & \textbf{1} & \textbf{1}  \\
\hline
\end{tabular}
}
\end{table*}

\subsection{Quantitative Comparison with the State-of-the-arts}
\textbf{Evaluation on ASR and query times:} Recall that the proposed DTA aims to lower the query times while maintaining a pleasing attack success rate. The success rate with sufficiently high query times may reach a certain bound, but this is not the scope of the current work. Hence, we make the comparison under a set of limited queries by setting the maximal number of queries to 100, 200, 300, 400, and 500. The selected competitors are all hard-label attacks. Thus, an attack is successful only within the predefined query number and otherwise, failure occurs. The comparisons on CIFAR-10 under $ \epsilon = 8 $ and $ \epsilon = 16 $ are shown in Table \ref{tab:cifar_asr}, while the results on SVHN are listed in Table \ref{tab:svhn_asr}. It can be seen that DTA achieves higher attack success rates than the competitors in most cases, which validates that the proposed generative model can synthesize effective adversarial examples. It should be especially noted that the average query number required by DTA is much smaller than that required by the other methods. 

The experiment on ImageNet poses a challenging case for our end-to-end adversarial example generation since the data is much more complex than CIFAR-10 and SVHN. The results on ImageNet are listed in Table \ref{tab:imagenet}, where we consider the perturbation level of $ \epsilon =16 $. The maximal number of queries is limited to 100, 200, 300, 400, and 500. We see that our method is superior to baselines on all metrics and is competitive with Rays on ASR in most instances. Again, DTA requires a very limited number of queries to perform a successful attack. Considering the attack performance in query-limited scenarios, we report the empirical results of Bandits, Rays, TA, and CGBA in the following sections. Besides, we prefer to report the results under the noise budget $\epsilon =16$ in most cases.

\textbf{Evaluation on Defense Model:} To evaluate the performance of the DTA on attacking robust models, we make a comparison by employing adv-inception-v3 (Adv-Inc-v3) \cite{iclr/TramerKPGBM18}, Ens3-adv-inception-v3 (Inc-v3{$_{ens3}$}) \cite{iclr/TramerKPGBM18}, Ens4-adv-inception-v3 (Inc-V3{$_{ens4}$}) \cite{iclr/TramerKPGBM18}, and Ens-adv-inception-resnet-v2 (IncRes-v2{$_{ens}$}) \cite{iclr/TramerKPGBM18} as the target models, all of which are adversarially trained. We first employ the selected 1000 images mentioned above to generate corresponding adversarial images on VGG-16 and then to test these generated examples' attack performance on these four defense models. All these pre-trained models' parameters can be available from the GitHub repository tf\_to\_pytorch\_model\footnote{https://github.com/ylhz/tf\_to\_pytorch\_model}. The results illustrated in Table \ref{tab:tab_eval_defense} show that DTA has achieved about 11.32$\sim$15.77\% attack success rate on these robust models. The baseline methods, Bandits, Rays, TA and CGBA, however, can only obtain 5.74$\sim$13.65\%, 3.14$\sim$7.79\%, 2.88$\sim$5.19\% and 0.77$\sim$3.44\%, respectively. It implies that the adversarial examples generated by DTA are more prone to attack deep models successfully, even on defense models.

\begin{table}
    \centering
    
    \caption{Evaluation adversarial robust accuracy (lower is better, $\downarrow$) on defense models. We first report the clean accuracy of the selected 1000 images, and the following results come from the generated adversarial examples by each attack method.}
    \label{tab:tab_eval_defense}
    \small
    \renewcommand{\arraystretch}{1.0}
    \setlength\tabcolsep{0.5pt}    
    \begin{tabular}{ccccc} 
        \toprule
        Method  & Adv-Inc-v3     & Inc-v3$_{ens3}$ & Inc-v3$_{ens4}$  & IncRes-v2$_{ens}$  \\ 
        \midrule
        Clean   & 94.80          & 93.20         & 91.30          & 97.40            \\ 
        \hdashline
        Bandits & 86.15          & 87.5          & 88.16          & 92.69            \\
        Rays    & 87.01          & 82.91         & 83.03          & 91.66            \\
        TA      & 89.61          & 90.32         & 86.94          & 92.63            \\
        CGBA    & 91.34          & 92.43         & 88.22          & 96.25            \\
        Ours    & \textbf{79.03} & \textbf{79.4} & \textbf{79.98} & \textbf{83.39}   \\
        \bottomrule
    \end{tabular}
\end{table}


\begin{table}[tp]
	\caption{Evaluation of attacks by DEEPSEC. Where the MR, ACAC, ACTC and PSD are lower the better ($\downarrow$), and the NTE, RGB and RIC are higher the better ($\uparrow$).}
	\label{tab:tab_eval}
	\centering
    \small
	\renewcommand{\arraystretch}{1}
    \setlength\tabcolsep{0.5pt}
	\begin{tabular}{cccccccc} 
        \toprule
        Method       & MR(\%)         & ACAC          & ACTC          & PSD             & NTE           & RGB           & RIC            \\ 
        \midrule
        Bandits       & 31.34          & 0.53          & 0.44          & 197.6           & 0.27          & 0.33          & 0.32           \\
        Rays          & 27.81          & 0.65          & 0.30          & 182.17          & 0.33          & \textbf{0.52} & \textbf{0.63}  \\
        TA            & 34.62          & 0.61          & 0.27          & 167.25          & 0.41          & 0.46          & 0.54           \\
        CGBA          & 12.67          & 0.58          & 0.38          & 164.97          & 0.45          & 0.47          & 0.49           \\
        Ours & \textbf{45.98} & \textbf{0.73} & \textbf{0.19} & \textbf{153.63} & \textbf{0.51} & 0.33         & 0.45           \\
        \bottomrule
    \end{tabular}
\end{table}

Besides the above comparisons, we are also interested in evaluating the performance of our method and the competitors by using different metrics. DEEPSEC \cite{sp/LingJZWWLW19} is a useful tool for the assessment of adversarial examples, which provides ten evaluation indicators. Specifically, from the perspective of classification outcomes, DEEPSEC provides 1) Misclassification Ratio (MR), 2) Average Confidence of Adversarial Class (ACAC), and 3) Average Confidence of True Class (ACTC). From the perspective of imperceptibility, DEEPSEC provides 1) Average $L_p$ Distortion $ ALD_{p} $, including $ L_{0} $, $ L_{2} $, and $ L_{\infty} $, 2) Average Structural Similarity (ASS), and 3) Perturbation Sensitivity Distance (PSD). From the perspective of the robustness of adversarial samples, DEEPSEC provides 1) Noise Tolerance Estimation (NTE), 2) Robustness to Gaussian Blur (RGB), 3) Robustness to Image Compression (RIC), and 4) Computation Cost (CC). We select 7 indicators as the evaluation metrics, as shown in Table \ref{tab:tab_eval}. In this experiment, we optimize the ResNet-20 model \cite{cvpr/HeZRS16} on CIFAR-10 until the best performance ($\geq 90\%$) on the test set is obtained. Then, 1000 images are selected as the normal examples by DEEPSEC (according to the given instructions). The adversarial examples are generated by Bandits, Rays, TA, CGBA, and DTA. The maximal query number is set to 100. The target model is ResNet-20. Given all generated adversarial examples during the attack, we finally employ DEEPSEC to compute the corresponding metrics. As shown in Table \ref{tab:tab_eval}, our method is superior to other methods in terms of misclassification rate and robustness by MR (45.98\%), ACAC (0.73), ACTC (0.19), PSD (153.63) and NTE (0.51), which reveals that the adversarial examples generated by DTA have stronger attack capabilities and anti-detection capabilities.

\subsection{Query Distribution}
\begin{figure}[htp]
    \setlength{\abovecaptionskip}{0.1cm}
    \setlength{\belowcaptionskip}{-10pt} 
	\centering
	\includegraphics[width=0.47\textwidth]{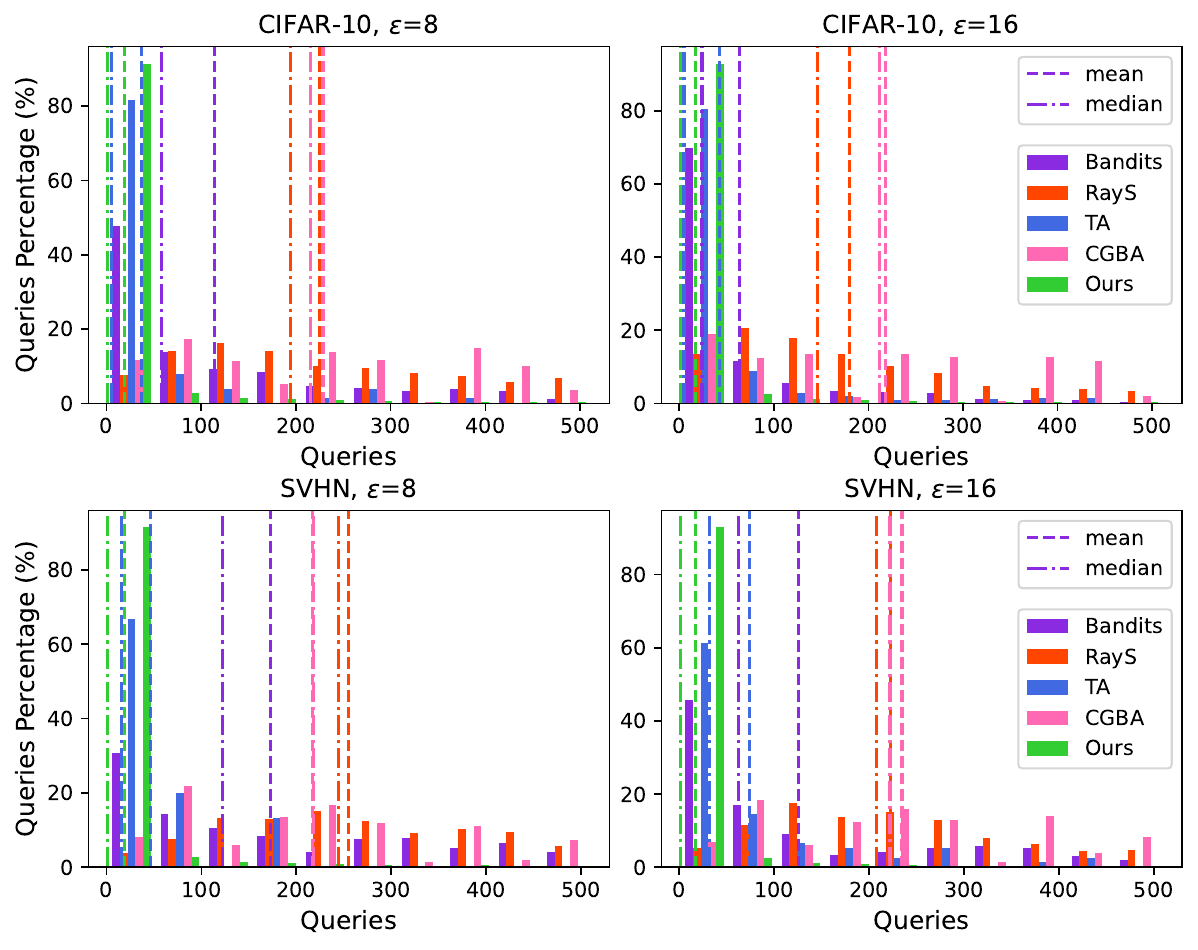}
	\caption{Histogram of the query number that is required to perform a successful attack on CIFAR-10 and SVHN. The median and mean lines denote the median query count and the average query count of the corresponding method (indicated by color), respectively.}
	\label{fig:his}
\end{figure}

To see the advantage of the proposed framework on query number for each attack, we plot the histogram of query numbers used to perform a successful attack in Fig. \ref{fig:his} for CIFAR-10 and SVHN. The test sets of CIFAR-10 and SVHN are used to compute the statistics, while ShuffleNetV2 \cite{eccv/MaZZS18} is employed as the target model. The maximal query number is limited to 500. For clearance, each bar denotes how many normal examples yield successful attacks with the times as noted in the x-axis. As observed, in all cases, the proposed DTA can perform a successful attack based on most examples with only ONE time. The average counts of query times by DTA for CIFAR-10 and SVHN are only 17.05 and 38.08 under $ \epsilon=16 $, respectively. Notably, on ShuffleNetV2, DTA helps 88\% and 90\% examples to attack successfully within a handful of query times when $ \epsilon=8 $ and $ \epsilon=16 $, respectively. On the other hand, Rays and Bandits often require hundreds of queries to perform a successful attack, and a small number of queries (such as $\leq 100$) could not allow these methods to work well. As \cite{icml/GuoGYWW19} indicates, the distribution of the histogram is highly right-skewed and hence, the median query count is a more representative aggregate statistic than the average query count. The results show that the median values of our method are only ONE in all cases on CIFAR-10, which sufficiently validates the proposed generative idea on generating adversarial examples.

\begin{figure}[htp]
    \setlength{\abovecaptionskip}{0.1cm}
    \setlength{\belowcaptionskip}{-5pt} 
	\centering
	\includegraphics[width=0.47\textwidth]{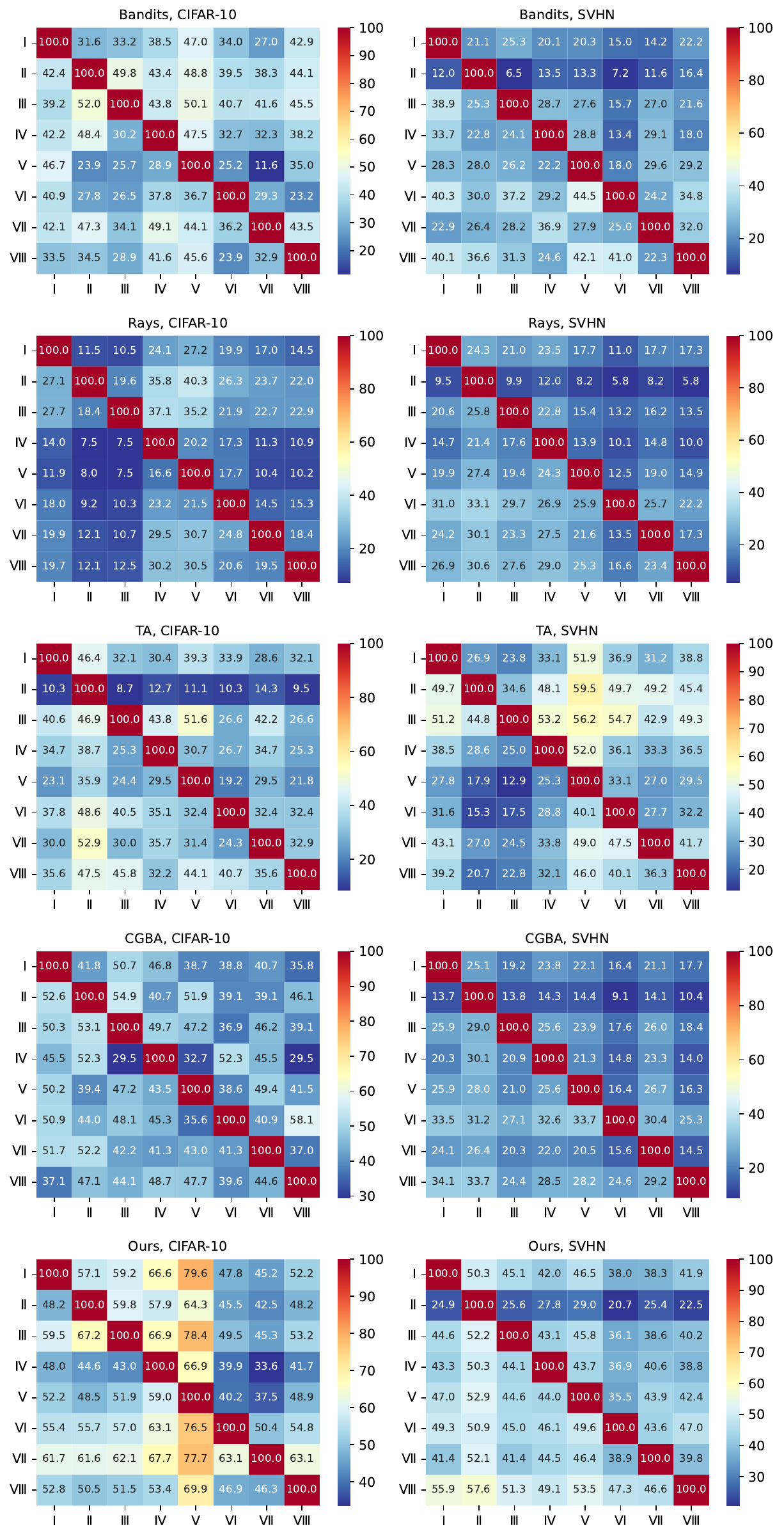}
	\caption{The attack success rate matrix of Bandits, Rays, TA and DTA on CIFAR-10 (left) and SVHN (right).}
	\label{fig:confusion}
\end{figure}

\subsection{Transferability}
The motivation of the current work states that the generation of adversarial examples is generally based on the assumption of transferability, which is saying that an adversarial example generated according to a model can be used to attack the other different models. To see this assumption is valid for black-box attacks, here, we follow the previous work \cite{aaai/ZhaoCWL20,nips/DolatabadiEL20} and examine the transferability of the generated adversarial examples across different models on CIFAR-10 and SVHN. Specifically, we select 8 models including ResNet-50 \cite{cvpr/HeZRS16}, VGG-16 \cite{corr/SimonyanZ14a}, VGG-19 \cite{corr/SimonyanZ14a}, ShuffleNetV2 \cite{eccv/MaZZS18}, MobileNetV2 \cite{corr/abs-1801-04381}, InceptionV3 \cite{cvpr/SzegedyVISW16}, DenseNet-169 \cite{cvpr/HuangLMW17}, and GoogLeNet \cite{cvpr/SzegedyLJSRAEVR15}. Following the settings in \cite{iclr/KurakinGB17a}, we randomly select 1000 images from the test set, which are classified correctly by the model whereas the corresponding adversarial examples are misclassified. The generated adversarial examples are used to attack the other models. For a fair comparison, we set $ \epsilon=16 $ and the maximal query number to 500 for all cases.

DTA is compared with Bandits, Rays and TA in the untargeted black-box attack settings. The ASR matrix on the two datasets is shown in Fig. \ref{fig:confusion}. The row represents which model is targeted during the generation of adversarial examples (we only preserve these adversarial examples that can attack the target model successfully), while the column represents which model is attacked by the aforementioned generated examples. From this figure, we can see that the transferability ASR on CIFAR-10 of DTA is from 33.6\% to 79.6\%, while the baseline methods are 11.6\% to 52.0\%, 7.5\% to 40.3\% and 8.7\% to 52.9\%, respectively. It means that the examples generated by DTA produce a higher(about 26.1\%$\sim$27.0\% higher in most cases) attack success rate on changed models than those by Bandits and Rays, validating the superior transferability of DTA. This is because baseline methods heavily rely on the feedback of the target model during each query and cannot extract transferable features. By contrast, our method learns the adversarial distribution that does not collapse to a certain model.

\subsection{Dataset- and Model-Agnostic Attack}
To evaluate the performance of DTA on the examples with different semantics and model structures, we first conduct the attack experiments on other datasets than the training ImageNet dataset. Specifically, the test datasets include VOC 2007 (VOC-07) \cite{Everingham10}, VOC 2012  (VOC-12) \cite{Everingham10}, Plasces365 (Pla-365) \cite{zhou2017places}, and Caltech101 (Cal-101) \cite{cvpr/LiFP04}. The target models include VGG-19 \cite{corr/SimonyanZ14a}, InceptionV3 \cite{cvpr/SzegedyVISW16}, ResNet-152 \cite{cvpr/HeZRS16}, and WideResNet-50 \cite{bmvc/ZagoruykoK16}, all of which are implemented in PyTorch. The attack results are illustrated in Table \ref{tab:cross_data}, which shows that the DTA trained on ImageNet is available to generate effective adversarial examples on other datasets without retraining. In certain situations, the attack success rate can exceed 90\%, where the maximal query size is limited to 100. To be clear, we do not care about how the ground truth labels of those datasets affect the current DTA, but only calculate the attack success rate by comparing the outputs of the original clean image and the corresponding adversarial counterpart, just as the Evasion Rate \cite{corr/abs-2012-06024}.

\begin{table}
    \centering
    \caption{The attack success rates on other datasets that are not involved in the DTA training progress. Here we report the Evasion Rate \cite{corr/abs-2012-06024} of each dataset on the victim models, which are pre-trained on ImageNet.}
    \label{tab:cross_data}
    \small
    \setlength{\tabcolsep}{3pt}
    \begin{tabular}{cccccc} 
    \toprule
     Metric & VOC-07 & VOC-12 & Pla-365 & Cal-101   \\ 
    \midrule
    VGG-19                & 91.7    & 93      & 90.9       & 93.5              \\
    InceptionV3           & 87.5    & 90.8    & 91.1       & 93.6              \\
    ResNet-152            & 85.1    & 89.2    & 87.3       & 94.4            \\
    WideResNet-50         & 86.1    & 89.7    & 84.1       & 93.4              \\
    \bottomrule
    \end{tabular}
\end{table}

We further apply our DTA to attack transformers, which are pretty different from traditional CNN, including ViT-16 \cite{iclr/DosovitskiyB0WZ21}, ViT-32 \cite{iclr/DosovitskiyB0WZ21}, and Swin-B \cite{iccv/LiuL00W0LG21}. Where the DTA is trained on the collected data pairs from CNNs with noise budget $\epsilon=16$, the empirical results we report in Table \ref{tab:cross_model} show that DTA can obtain 26\%$\sim$41\% attack success rate with limited queries. This phenomenon demonstrates that even in attacking transformers, DTA can still generate adversarial examples and achieve an acceptable attack effect on different ViT models. Furthermore, it illustrates the high adaptability of DTA in model-agnostic black-box scenarios.

\begin{table}
\centering
\caption{The attack success rate (ASR, \%) v.s. average queries (Avg.Q) of different transformers trained on the ImageNet dataset.}
\small
\renewcommand{\arraystretch}{1.0}
\setlength\tabcolsep{4pt}
\label{tab:cross_model}
    \begin{tabular}{ccccccc} 
        \toprule
        Metric                 & Model   & 100   & 200   & 300   & 400   & 500    \\ 
        \midrule
        \multirow{3}{*}{ASR}   & ViT-16  & 29.93 & 33.04 & 35.16 & 35.91 & 37.53  \\
                               & ViT-32  & 34.35 & 38.08 & 39.28 & 40.61 & 41.01  \\
                               & Swin-B & 26.63 & 29.35 & 29.23 & 30.77 & 31.48  \\ 
        \midrule
        \multirow{3}{*}{Avg.Q} & ViT-16  & 15.59 & 2.89  & 40.50 & 49.03 & 60.91  \\
                               & ViT-32  & 13.79 & 24.58 & 34.24 & 41.51 & 43.57  \\
                               & Swin-B & 17.00 & 29.83 & 36.35 & 48.08 & 56.22  \\
        \bottomrule
    \end{tabular}
\end{table}

\subsection{Ablation Study}
\subsubsection{Loss and Hyper-parameters}
The proposed method concerns the settings of the MSE loss and the hyper-parameters, such as $L$ and $K$, which affect the model depth. We examine the influence of these factors on the CIFAR-10 dataset. The target model is the pre-trained VGG-16 \cite{corr/SimonyanZ14a}. During the attack, the maximal number of queries is limited to 500. 

\begin{table}[htp]
    \caption{The attack performance comparison of DTA optimized with and without the MSE loss, w. means with MSE loss, w.o. means without MSE loss.}
    \label{tab:mse}
    \centering
    \renewcommand{\arraystretch}{1.0}
    \setlength\tabcolsep{1.0pt}
    \small
    \begin{tabular}{cccccc}
        \toprule
        Metric         & \textbf{$\epsilon=8$ w. } & \textbf{$\epsilon=8$ w.o.} & \textbf{$\epsilon=16$ w. } & \textbf{$\epsilon=16$ w.o. } \\ 
        \midrule
        ASR(\%)           & 74.75                        & 62.31                          & 88.67                         & 81.79                           \\
        Avg. Q & 15.41                        & 11.69                          & 17.33                         & 16.39                           \\ 
        \bottomrule
    \end{tabular}
\end{table}

\begin{table}[ht]
    \caption{TThe attack success rate (ASR(\%)) and average query number (Avg.Q) under different $ K $'s, here we fix the flow block $L$ as $ L=3 $.}
    \label{tab:steps}
    \centering
    \renewcommand{\arraystretch}{1.0}
    \setlength\tabcolsep{10pt}
    \small
    \begin{tabular}{cccccc}
        \toprule
        Metric     & $K$=2 & $K$=4 & $K$=6 & $K$=8 \\ 
        \midrule
        ASR(\%)           & 81.81        & 81.44        & 81.87        & 81.65 \\
        Avg. Q & 13.64        & 14.97        & 14.71        & 14.46        \\ \bottomrule
    \end{tabular}
\end{table}

\begin{table}[htp]
    \caption{The attack success rate (ASR(\%)) and average query number (Avg.Q) under different $ L $'s, here we fix the flow steps $K$ as $ K=3 $.}
    \label{tab:levels}
    \centering
    \renewcommand{\arraystretch}{1.0}
    \setlength\tabcolsep{10pt}
    \small    
    \begin{tabular}{cccccc}
        \toprule
        Metric         & $L$=1 & $L$=2 & $L$=3 & $L$=4 \\
        \midrule
        ASR(\%)           & 81.40        & 81.98        & 81.82        & 81.77        \\ 
        Avg. Q & 12.86        & 14.17        & 13.54        & 14.57        \\ 
        \bottomrule
    \end{tabular}
\end{table}

First, we evaluate the performance of DTA with and without the MSE loss. If the MSE loss is not used, we mean that the updating step in the 4-th line of Alg. \ref{alg:alg1} is omitted. The comparison is listed in Table. \ref{tab:mse}, which shows that the flow model could benefit from the MSE loss, yielding notable improvement on both attack success rate and average query number.

Next, we test how model depth or representative capacity affects the attack performance. Two experiments are considered here. In the first one, we fix $L=3$ and examine the influence of $ K $ from $\{2, 4, 6, 8\}$. In the second one, we fix $K=2$ and evaluate the performance of $L$ from $\{1, 2, 3, 4\}$. The results are shown in Tables \ref{tab:steps} and \ref{tab:levels}, respectively. As seen, different settings produce a similar performance on both ASR and average query number, which suggests that the attack ability of the proposed model on CIFAR-10 does not benefit from the increasing of the model depth. This may be because the data in CIFAR-10 is simple and hence, we set $K=2$ and $L=2$ in small datasets, e.g., CIFAR-10 and SVHN. But in ImageNet, which contains complex data, we set $K=8$ and $L=5$.

\begin{figure}
    \centering
     \includegraphics[width=0.48\textwidth]{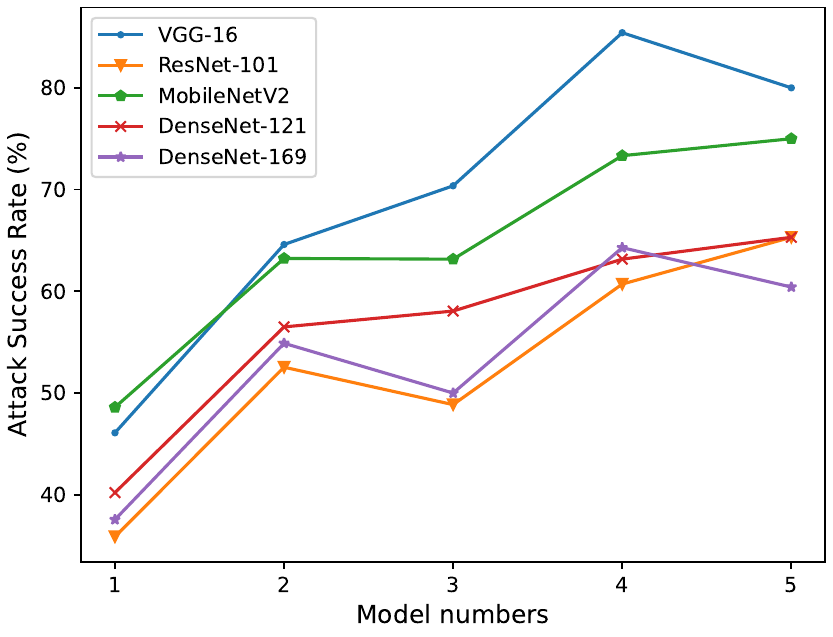}
    \caption{\small{The performance of DTA when using different numbers of attack models.}}
 
	\label{fig:ensemble}
\end{figure}

Furthermore, we examine the generalization ability of the adversarial examples generated by DTA, i.e., testing the performance of DTA by involving different numbers of white-box attack models. Specifically, we use ten pre-trained models on ImageNet, including VGG-19 \cite{corr/SimonyanZ14a},  ResNet-152 \cite{cvpr/HeZRS16}, InceptionV3 \cite{cvpr/SzegedyVISW16}, DenseNet-201 \cite{cvpr/HuangLMW17}, WideResNet-50 \cite{bmvc/ZagoruykoK16}, VGG-16 \cite{corr/SimonyanZ14a}, ResNet-101 \cite{cvpr/HeZRS16}, MobileNetV2 \cite{corr/abs-1801-04381}, DenseNet-121 \cite{cvpr/HuangLMW17}, and DenseNet-169 \cite{cvpr/HuangLMW17}. The first five models are used for generating the training adversarial examples, while the rest are used as the black-box test models to evaluate the attack effect of DTA. In this experiment, we select different numbers of models from the first five ones for example generation, where the results are plotted in Fig. \ref{fig:ensemble}. It can be clearly observed that the more models used in performing the attack, the better performance can be obtained on all the target models. This indicates that we can use more models in the process of sample adversarial examples to gain an increased universal success rate on black-box attacks.

\subsubsection{Improved performance by shifted means and stds}
It is common to sample from a Gaussian distribution during the inverse of a normalized flow model to generate the expected data; however, in this work, we found that if we simply sample from $\mathcal{N}(0,1)$ to obtain the adversarial examples, the attack performance will behave badly. Instead, for the well-trained normalized flow model, we first input the training data to obtain its corresponding latent space $\bm{z}$, then count the mean $\hat{\bm{\mu}}$ and variance $\hat{\delta}$ of $\bm{z}$ and use it as the Gaussian distribution $\hat{\mathcal{N}}(\hat{\bm{\mu}},\hat{\delta}^2)$ as the adversarial latent space $\hat{\bm{z}}$ for sampling to enhance the attack performance. We report the empirical results in Table. \ref{tab:shifted}, as the results show, equipment the shifted mean $\hat{\bm{\mu}}$ and std $\hat{\delta}$, the attack success rate are improved by 49.16\% $\sim$ 59.65\%, 35.35\% $\sim$ 36.72\% and 26.38\% $\sim$ 37.49\% over CIRAR-10, SVHN and ImageNet, respectively. These results have demonstrated well that the latent space has been shifted guided by the adversarial-clean example pairs.

\begin{table}
    \centering
    \caption{Attack success rate of with (\textbf{w.}) and without (\textbf{w.o.}) shifted mean and std on three benchmark datasets under noise budget $\epsilon=8$ and $\epsilon=16$.}
    \label{tab:shifted}
    \small
    \begin{tabular}{c|cc|cc|cc} 
        \hline
           \multirow{2}{*}{$\epsilon$}  & \multicolumn{2}{c|}{CIFAR-10} & \multicolumn{2}{c|}{SVHN} & \multicolumn{2}{c}{ImageNet}  \\
             & \textbf{w.o.}  & \textbf{w.}                    & \textbf{w.o.}  & \textbf{w.}                & \textbf{w.o.}  & \textbf{w.}                    \\ 
        \hline
        8  & 13.92 & 73.57                 & 18.64 & 55.36             & 17.83 & 44.21                 \\
        16 & 31.33 & 80.49                 & 48.76 & 84.11             & 40.19 & 71.68                 \\
        \hline
    \end{tabular}
\end{table}

\subsubsection{Compare with GAN-based}
\begin{figure}
    \centering
     \includegraphics[width=0.48\textwidth]{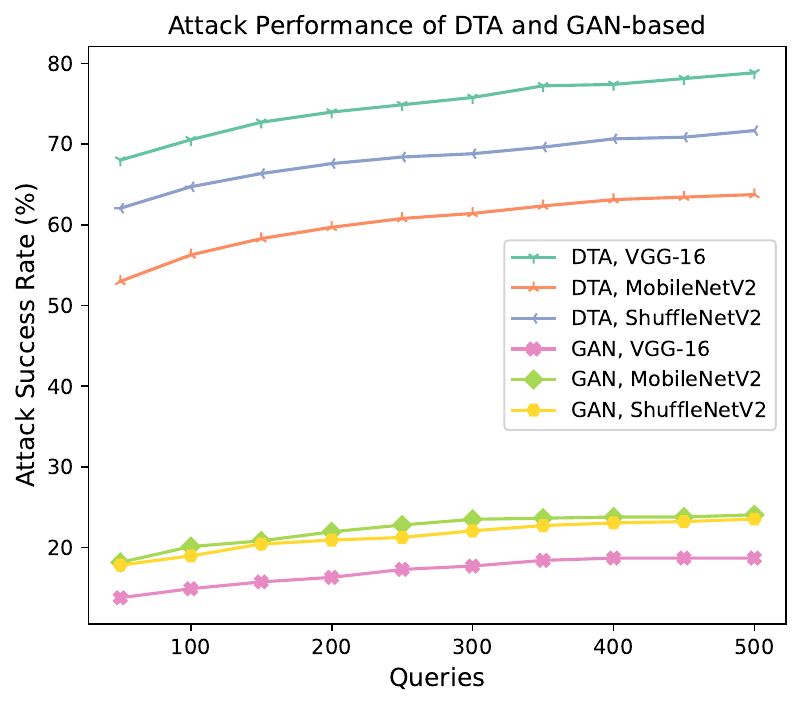}
    \caption{\small{The performance of DTA when using different numbers of attack models. Where the dataset is ImageNet and the noise budget $\epsilon=16$.}}
 
	\label{fig:compare_vs_gan}
\end{figure}

As we declared above, the adversarial and normal examples come from different distributions, which are misaligned but transferable, and we characterize the adversarial distribution with locally collected adversarial examples in a generative manner; more specifically, a conditional normalized flow is involved in learning the transformation in this paper. To verify whether other generative models are qualified for this task well or not, we apply the whole same pipeline to the GAN borrowed from GAP \cite{cvpr/PoursaeedKGB18} and present the attack performance in Fig. \ref{fig:compare_vs_gan}. As the results show, the GAN can also learn the mapping relationship between these two types of samples, but its attack capability is unsatisfactory; in addition, as the query budget increases, the attack success rate of DTA will increase significantly, while the GAN-based method will not. Again, these results demonstrate the superiority of our proposed conditional likelihood-based DTA method in generating examples belonging to adversarial distributions.

\section{Conclusions}
\label{Sec:conclusion}
In this paper, we propose a novel hard-label black-box adversarial attack framework based on a generative idea. The motivation states that the public datasets enforce the public models to learn a common distribution, causing that the models exhibit similar vulnerability. Hence, the adversarial distributions of different models could also be similar, which inspires the transferability assumption in many adversarial attack methods. Based on such an assumption, we advocate that there could be a certain mapping from the distribution of normal examples to the distribution of adversarial examples. Along with this, a conditional normalizing flow-based generative model is developed to implement the mapping function. We can optimize the flow model to explicitly correlate the adversarial examples with Gaussian-style hidden representations by collecting a batch of adversarial examples from the existing white-box attacks. To diversify the generation process, the normal examples are fed into the conditions of the probabilistic model. An elaborated generation process helps us to improve the performance of the generated examples. Extensive experiments validate the proposed idea and demonstrate the superiority of DTA on attack success rate, average query number and median query number. Especially, our method can achieve a successful attack within only ONE query, which verifies that we have learned the adversarial distribution. By contrast, the other hard-label methods generally require hundreds of queries to accomplish an attack. We also surprisingly find that the proposed model can perform effective cross-dataset attacks, which means that the model is not sensitive to the label space of the classification task. In summary, this work provides a promising framework with the advantages of low query times, high success rate, and an efficient inference process, which could guide future research on adversarial attacks in a new direction.

\section*{Declarations}
\begin{itemize}
\item \textbf{Availability of data and material: }The data that support the findings of this study are openly available in at https://www.cs.toronto.edu/~kriz/cifar.html, http://ufldl.stanford.edu/housenumbers/, and https://image-net.org/, reference number [51], [54], and [44].

\item \textbf{Funding: }This work is supported in part by the National Natural Science Foundation of China under Grant 62162067, 62101480 and 62362068, Research and Application of Object Detection based on Artificial Intelligence, in part by the Yunnan Province expert workstations under Grant 202305AF150078 and the Scientific Research Fund Project of Yunnan Provincial Education Department under 2023Y0249.

\item  \textbf{Acknowledgements: }Not applicable

\end{itemize}

\bibliography{sn-bibliography}

\end{document}